\documentclass[journal]{IEEEtran}
\ifCLASSINFOpdf
\else
\fi
\usepackage{amsmath}
\usepackage[ruled]{algorithm2e}
\usepackage{multirow}
\usepackage{graphicx} %
\usepackage{subfigure}
\usepackage{color}
\usepackage{cite}
\usepackage{xcolor}
\usepackage{makecell}
\usepackage{cleveref}
\usepackage{caption}

\begin{document}
%
\title{Learning to Purification for Unsupervised Person Re-identification}
\author{
    \IEEEauthorblockN{Long Lan$^{*}$, \IEEEmembership{Member, IEEE}, Xiao Teng$^{*}$, Jing Zhang, \IEEEmembership{Member, IEEE}, Xiang Zhang, \IEEEmembership{Member, IEEE},   \\and Dacheng Tao, \IEEEmembership{Fellow, IEEE}}
    
    }

\maketitle

\begin{abstract}
Unsupervised person re-identification is a challenging and promising task in computer vision. Nowadays unsupervised person re-identification methods have achieved great progress by training with pseudo labels. 
However, how to purify feature and label noise is less explicitly studied in the unsupervised manner.
To purify the feature, we take into account two types of additional features from different local views to enrich the feature representation. The proposed multi-view features are carefully integrated into our cluster contrast learning to leverage more discriminative cues that the global feature easily ignored and biased.
To purify the label noise, we propose to take advantage of the knowledge of teacher model in an offline scheme. Specifically, we first train a teacher model from noisy pseudo labels, and then use the teacher model to guide the learning of our student model. In our setting, the student model could converge fast with the supervision of the teacher model thus reduce the interference of noisy labels as the teacher model greatly suffered.
After carefully handling the noise and bias in the feature learning, our purification modules are proven to be very effective for unsupervised person re-identification.
Extensive experiments on three popular person re-identification datasets demonstrate the superiority of our method. Especially, our approach achieves a state-of-the-art accuracy 85.8\% @mAP and 94.5\% @Rank-1 on the challenging Market-1501 benchmark with ResNet-50 under the fully unsupervised setting. The code will be released.
\end{abstract}


%
\IEEEpeerreviewmaketitle

\section{Introduction}
%
%
%
%
\IEEEPARstart{P}{erson} re-identification (ReID) aims to retrieve the same person under different camera views. It has attracted widespread attentions in the computer vision community due to its great potential in real world applications~\cite{zhang2020empowering}. Although great performance has been achieved in the supervised person ReID setting, the demand of human annotation heavily limits the application. To make it more scalable in the real world, the task of unsupervised person ReID has been raised and attracted increasing more attention as it requires no human annotation. 

Unsupervised person ReID mainly includes two categories, unsupervised domain adaptation (UDA) person ReID and purely unsupervised learning (USL) person ReID \cite{hu2021hard}. The UDA methods aim to learn from the annotated source dataset and transfer the knowledge to the unlabeled target dataset \cite{kumar2020unsupervised, ge2020structured, ge2019mutual}. They usually adopt the two-stage training strategy. At first the model is pre-trained on the labeled source dataset, then the unlabeled target dataset is utilized to finetune the model. Unlike existing general unsupervised domain adaptation setting where the source domain and target domain share the same label space \cite{wei2021metaalign, xiao2021dynamic, na2021fixbi}, UDA person ReID usually assumes there are no interactions between source and target domains in the label space, thus compared with existing unsupervised domain adaptation setting~\cite{zhang2019category,gao2021dsp,wang2021exploring}, UDA person ReID is more challenging. Similar to other general unsupervised domain adaptation methods, UDA ReID methods are also proposed based on the assumption that the discrepancy between source domain and target domain is not significant, and the performance of these methods will drop significantly when the gap between source and target domains is large.

To further relax the dependency on labeled source dataset, the USL methods directly learn from the unlabeled target dataset, which require no annotation information from other domains \cite{lin2019bottom, lin2020unsupervised, zeng2020hierarchical, zeng2020energy}. Thus, compared with the UDA person ReID, the USL person ReID is more scalable. Nowadays state-of-the-art USL person ReID methods have achieved great 
progress by training the model with the pseudo labels generated by clustering algorithm \cite{ge2020self, dai2021cluster, hu2021hard}. These methods hold the assumption that the images of the same person share higher similarity in the feature space, thus will be more likely to be collected in the same cluster. Generally, these methods can be regarded as two-stage training schemes, firstly a clustering algorithm is applied to divide the features of images into different clusters, and assign pseudo labels to different clusters accordingly. Then the model is trained with generated pseudo labels. These two stages are conducted in an iterative scheme as they can promote each other in the whole training process. Based on this training framework, memory-based contrastive learning methods have achieved the state-of-the-art performance nowadays by taking advantage of contrastive learning with image features stored in the memory bank \cite{ge2020self, dai2021cluster, hu2021hard}.

\begin{figure*}[htbp]
\centering
\subfigure[Cluster contrast learning]{\includegraphics[height=4.60857cm,width=7.55cm]{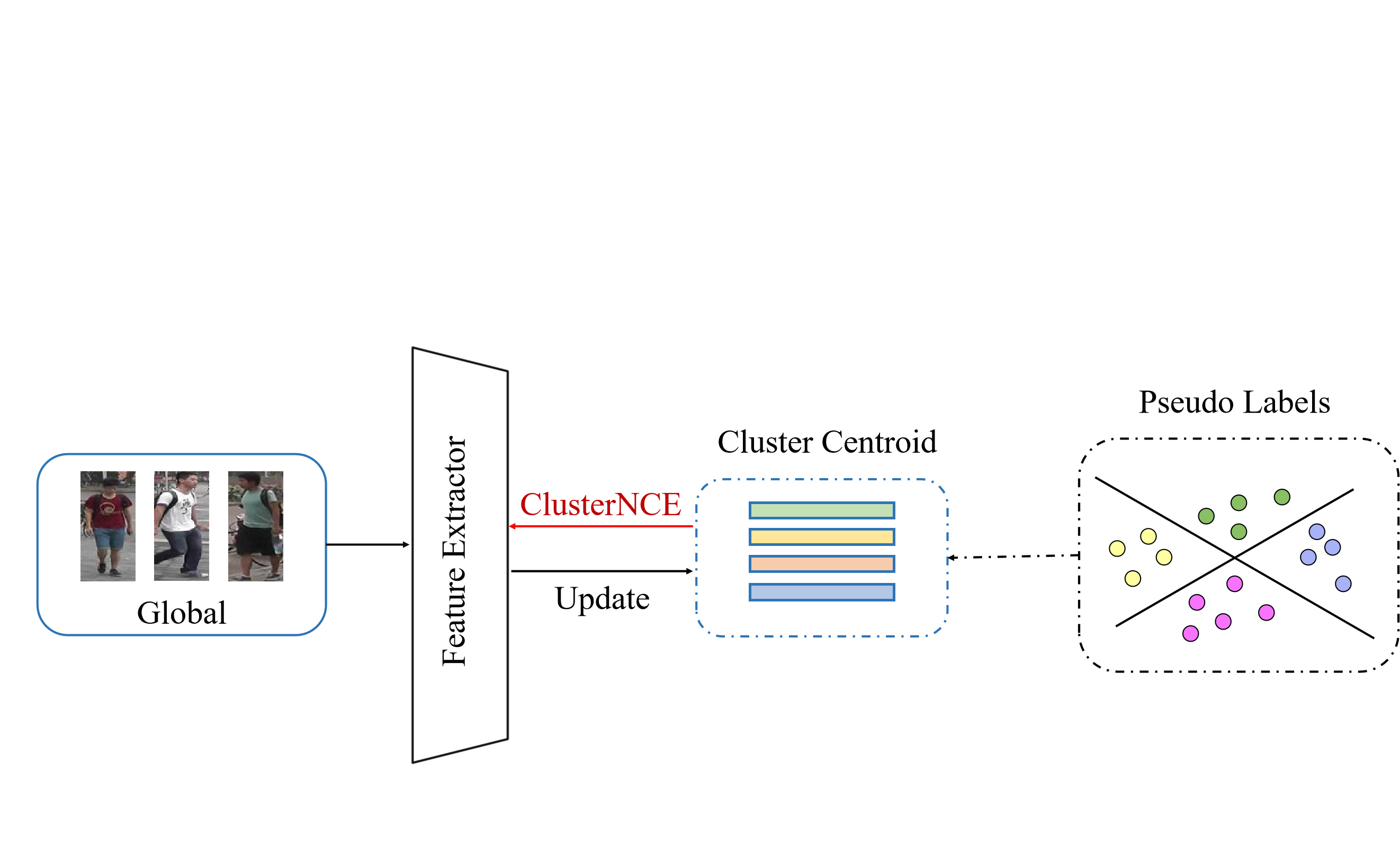}}\hspace{6mm}
\subfigure[Cluster contrast learning with purification modules]{\includegraphics[height=4.62cm,width=9.563cm]{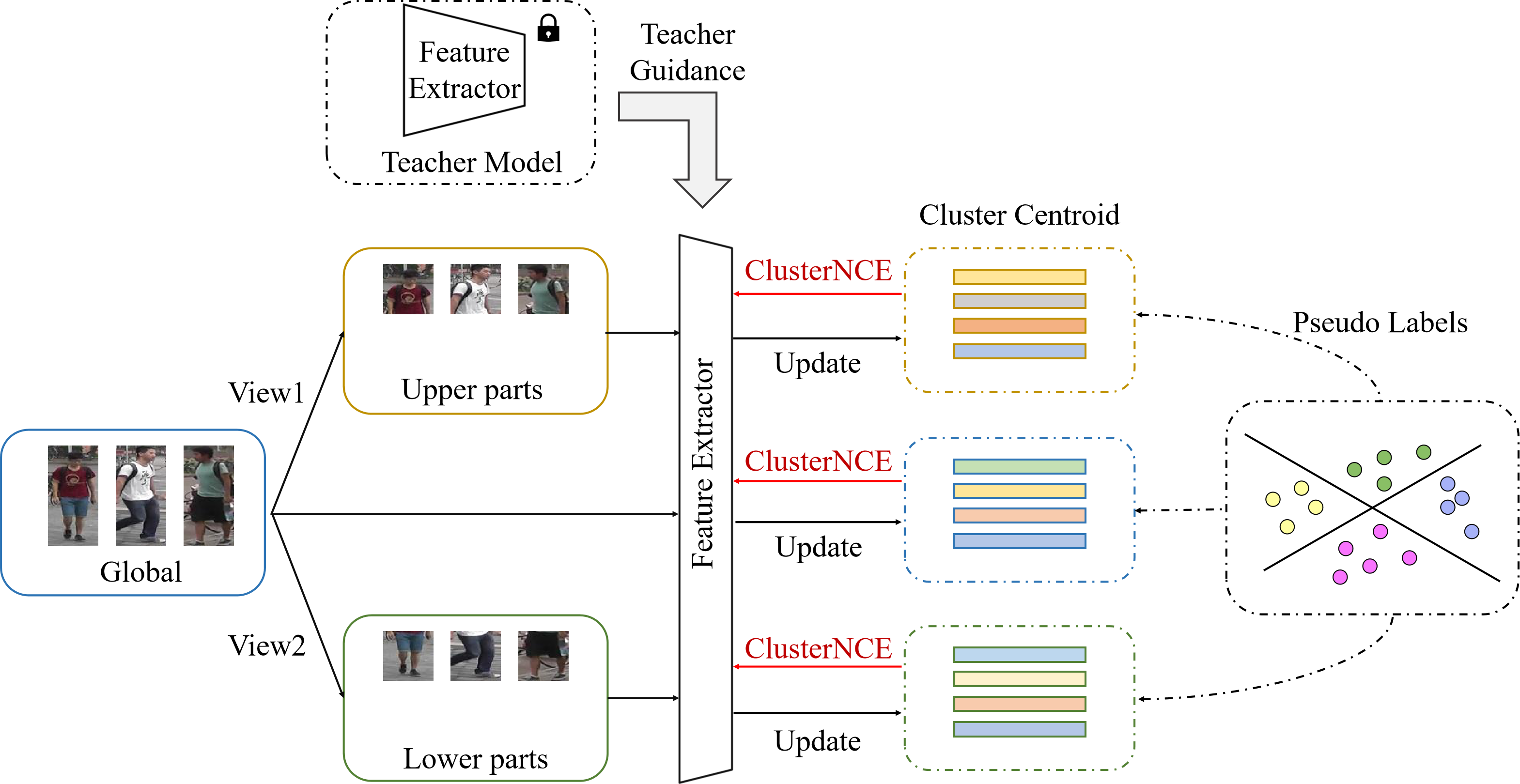}}
\caption{ Comparison of cluster contrast learning with/without purification modules. (a) Cluster contrast learning trains the model solely with cluster centers of global feature representations. (b) Besides the global level cluster contrast learning, feature and label noise purification modules are also applied. The former takes into account the features from two local views to purify the bias of the global feature involved. Meanwhile, the latter aims to purify the label noise by taking advantage of the knowledge of teacher model in an offline scheme.}
\label{fig3}
\end{figure*}

Although the above ReID methods have achieved great progress in recent years, the gap between supervised and unsupervised methods is still large. After carefully analysed the reason behind the phenomenon, we think the learning process of unsupervised person ReID is mainly influenced by the feature bias and label noise due to limited global feature representation power and lack of accurate predicted labels. As the aim of global feature learning is to capture the most salient clues of appearance to represent identities of different pedestrians, some non-salient but important detailed local cues can be easily ignored due to the limited scales and less diversities of the training dataset, which makes global features hard to distinguish from similar inter-class persons \cite{wang2018learning}. As a result, images with different identities but similar salient clues of appearance could be easily merged to the same cluster, which will make the learned feature representation biased. On the other hand, since the model is trained with pseudo labels generated by clustering algorithm, it will suffer from severe label noise during the whole convergence process as it is initialized with parameters pre-trained on ImageNet dataset, which has the significant discrepancy with person ReID datasets. To relieve the above problems, we propose the feature and label noise purification modules for unsupervised person ReID, as shown in Fig.~\ref{fig3}. 

Specifically, our method mainly includes two modules, the feature purification (FP) module and label noise purification (LP) module. The former takes into account the features from two local views to enrich the feature representation and purify the inherent feature bias of the global feature involved. Meanwhile, the latter aims to purify the label noise by taking advantage of the knowledge of teacher model in an offline scheme. As the noise will be inevitably introduced in the clustering process, the model will suffer from the label noise in the whole training process. Based on the phenomenon that the trained model is more accurate than the initialized model, intuitively the knowledge of the trained model can be utilized as the guidance to help the student model relieve the influence of noise in the training process. Our contributions can be concluded in the following:
\begin{enumerate}
\item[$\bullet$] We propose a feature purification module which carefully integrate the multi-view features in our cluster contrast learning framework and is proven to be effective in handling the bias of the global feature easily involved.

\item[$\bullet$] We further propose a label noise purification module which aims to relieve the label noise introduced by clustering procedure by taking advantage of the knowledge of teacher model. To our knowledge, this is the first work to apply offline knowledge distillation for unsupervised person ReID, and we find it is very effective for such task.

\item[$\bullet$] Extensive experiments are conducted on three popular person ReID benchmarks. The results show our method significantly outperforms existing state-of-the-art unsupervised person ReID methods. Specially, our method outperforms the state-of-the-art method \cite{dai2021cluster} by 3.2\%, 3.4\% and 6.2\% in terms of mAP on Market-1501, DukeMTMC-ReID, and MSMT17 datasets.
\end{enumerate}

\section{Related Work}
In this section, we introduce the most related work from three perspectives: 1) Unsupervised person ReID, which includes unsupervised domain adaptation (UDA) person ReID and purely unsupervised learning (USL) person ReID; 2) Part-based person ReID, which takes advantage of local parts of the person to get more discriminative feature representations; and 3) Knowledge distillation, which includes some techniques of knowledge distillation in different areas.

\subsection{Unsupervised Person ReID}
Unsupervised person ReID can be summarized into two categories, unsupervised domain adaptation (UDA) person ReID and purely unsupervised learning (USL) person ReID. The former aims to learn from the annotated source dataset and transfer the knowledge from the source domain to the unlabeled target domain \cite{kumar2020unsupervised, ge2020structured, ge2019mutual, hu2021hard}. While the latter directly trains on the unlabeled target dataset without any labeled data \cite{ge2020self, dai2021cluster, hu2021hard}. To make full use of unlabeled target dataset, unsupervised person ReID methods usually apply existing clustering algorithms, such as Kmeans \cite{macqueen1967some} and DBSCAN \cite{ester1996density} to generate pseudo labels for each sample in the target domain. Then the generated pseudo labels and the unlabeled dataset are used together to train the model in an iterative scheme \cite{fan2018unsupervised}. To further improve the quality of pseudo labels, many variants of pseudo label generation methods have been proposed. BUC \cite{lin2019bottom}  proposed a bottom-up clustering framework by exploiting the intrinsic diversity among identities and similarity within each identity to learn more discriminative feature representations. GLT \cite{zheng2021group} proposed a Group-aware Label Transfer algorithm that facilitates the online interaction and mutual promotion of the pseudo labels prediction and feature learning.  To avoid the label noise accumulation in a single model setting, MMT \cite{ge2019mutual} refines the noisy pseudo labels by optimizing two neural networks under the joint supervisions of off-line refined hard pseudo labels and on-line refined soft pseudo labels.

The above methods can be applied to both UDA ReID and USL ReID. However, different from USL ReID, UDA ReID also has the auxiliary labeled source dataset. Thus the key of UDA ReID methods is how to take advantage of labeled source dataset to improve the performance of the model on unlabeled target dataset. These methods usually work based on the assumption that the discrepancy between the source domain and the target domain is not significant and apply transfer learning techniques to tackle such problem. To further mitigate the gap between source and target domains, some domain-translation-based methods are proposed, which aim to take advantage of generative adversarial networks (GANs) \cite{goodfellow2014generative} to translate the source-domain images to have target-domain styles while preserving their original IDs \cite{deng2018image, wei2018person, ge2022structured}. Then the translated source dataset and unlabeled target dataset can be used together to train the model with some semi-supervised techniques. However, such methods highly depend on the performance of GANs as the quality of the translated dataset has a significant effect on the trained model. Instead of directly using GANs to translate the style of the source dataset to augment the target dataset, some works aim to regard the unsupervised person ReID problem as the multi-label classification problem by using labeled source dataset as reference persons \cite{yu2019unsupervised, wang2020unsupervised, li2022unsupervised}. These methods aim to learn the soft multi-label for each target-domain image by comparing them to the known auxiliary reference persons in the source domain. 

In this work we focus on USL ReID and our work is established on memory-based contrastive learning frameworks, which are the state-of-the-arts for unsupervised person ReID. SPCL \cite{ge2020self} proposed a self paced method which gradually create more reliable clusters to refine the hybrid memory and learning targets. To solve the problem of inconsistency in the memory updating process, CCL \cite{dai2021cluster} proposed a novel cluster contrast learning framework which was built on a cluster-level cluster memory dictionary and achieved great performance. In this work, to purify the feature and label noise for unsupervised person ReID, we also take into account local views and the knowledge of the teacher model. Thus we also discuss some works related to these techniques in the below. 

\begin{figure*}[thbp] 
\centering
\includegraphics[width=1.0\textwidth,height=7.5cm]{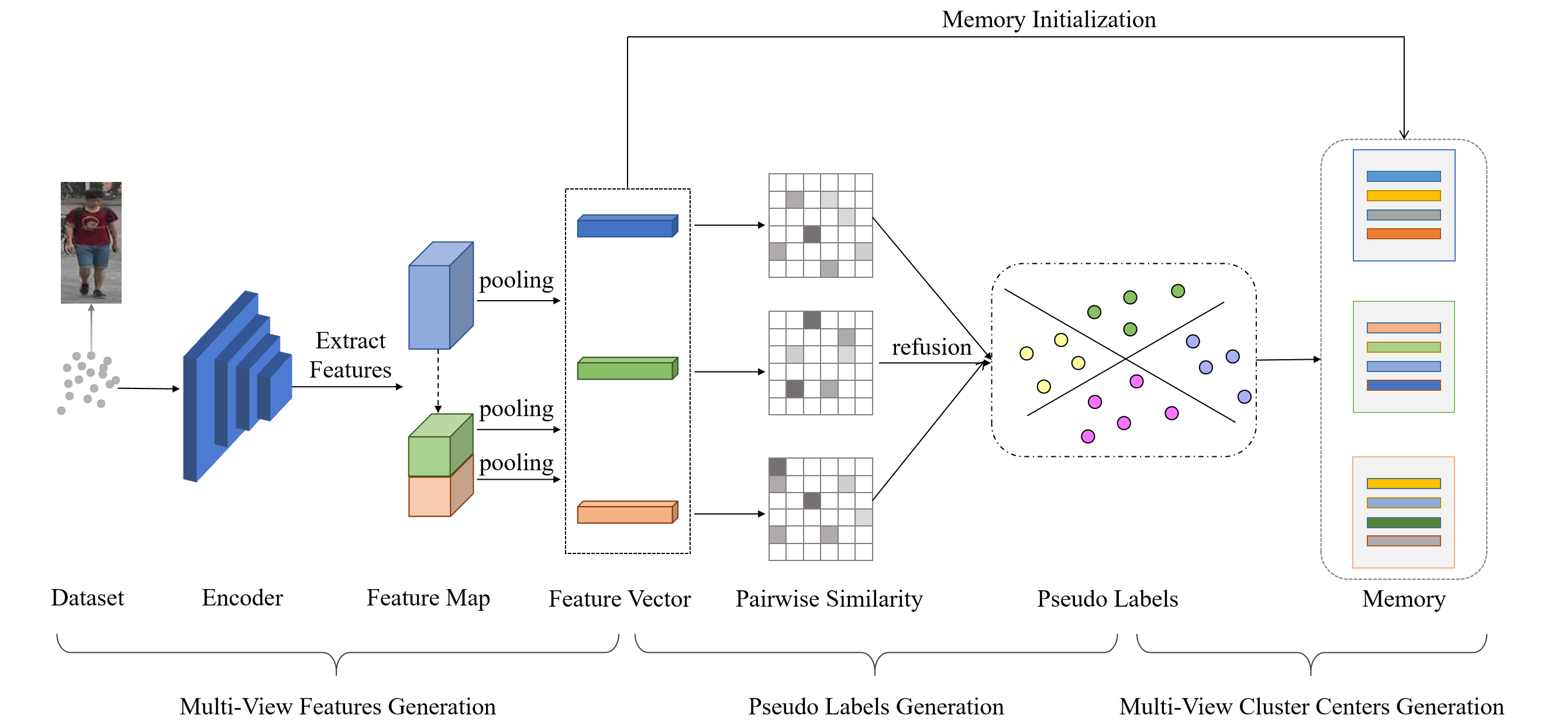}
\caption{Feature purification module. Besides the global feature, centers of local features are also maintained as independent cluster memory dictionaries. The DBSCAN clustering algorithm is applied in the fused pairwise similarity matrix to generate the consistent pseudo labels. The shared pseudo labels guide the local/global features to initialize their respective memory cluster representations. }\label{fig2a}
\end{figure*}

\subsection{Part-based Person ReID}
Most deep learning-based person ReID approaches take advantage of only the global feature of the person, which turns out to be sensitive to the missing key parts. To relieve the issue, recently many works focused on leveraging part discriminative feature representations. These works aim to make use of local parts to make more accurate retrieval. Part-based person ReID can be divided into three categories. In the first category, the prior knowledge like poses or body landmarks are required to be estimated to locate the accurate parts of the person. However, the performance of such approaches heavily rely on the accuracy of the pose or landmarks estimation models~\cite{zhang2021towards,xu2022vitpose}. The second category utilized the attention mechanism to adaptively locate the high activation in the feature map. But the selected regions lack of semantic interpretation \cite{fu2019horizontal}. The third category directly utilizes the predefined strips as it assumes the person is vertically aligned. Compared with the first category it is more scalable as it requires no extra pre-trained models, thus it is widely used in the person ReID and achieved great improvements in recent years. Specifically, PCB \cite{sun2018beyond} conducts uniform partition on the conv-layer for learning part-level discriminative fetaures. MGN \cite{wang2018learning} and RMGL\cite{wang2020receptive} utilize multi-Granularity parts to get a more accurate feature representation. These methods are all proposed for supervised person ReID. Similar to our work, SSG \cite{fu2019self} is also proposed for unsupervised person ReID and utilizes three sets of local and global features to represent persons. However, SSG generates pseudo labels by applying clustering algorithm on each set on them, which causes a huge cost of extra time computation. Furthermore, SSG updates the network with triplet loss and instance-level features, which may neglect high-level semantic meanings. Compared with SSG, our method uses the cluster center memory to capture the relation between local/global features and their own cluster centers, and our local and global branches share the same pseudo label set jointly generated by them to relieve the inherent feature bias in the single-scale global feature representations. 

\subsection{Knowledge Distillation}
The aim of knowledge distillation is to transfer the knowledge from the network to another. The original idea of knowledge distillation is to compress the knowledge from the teacher network to a smaller student network. Recently, more works have focused on self-knowledge distillation, which keeps the structure of the teacher and student network the same \cite{yun2020regularizing,gao2021dsp,kim2021self, li2021self}. These methods usually directly use the outputs of the teacher whose structure is the same as the student. Specifically, a simple but effective baseline was proposed for few shot learning in \cite{tian2020rethinking} by minimizing the loss where the target is the distribution of class probabilities induced by the teacher model. CS-KD \cite{yun2020regularizing} proposed a new regularization technique, which matches the distribution predicted between different samples of the same class. SSD \cite{li2021self} proposed a effective multi-stage training scheme for long-tailed recognition, which utilized the output of the teacher to generate soft label for the student. Similar to our work, a probability distillation module is proposed in \cite{cheng2022hybrid} which aims to align the probability distribution between the network and the teacher network updated by Exponential Moving Average (EMA) method. However, the teacher network updated in the online scheme is still limited in the feature representation and suffers from severe label noise~\cite{zhang2019category}. In the work, we aim to transfer the knowledge from the teacher to the student model in a multi-view and offline scheme directly in the feature space, and we find it is effective for unsupervised person ReID.

\section{Method}
\subsection{Cluster Contrast Learning Framework}
Let $X=\left\{x_{1}, x_{2}, \ldots, x_{N}\right\}$ denotes the unlabeled training set which contains $N$ instances. $F=\left\{f_{1}, f_{2}, \ldots, f_{N}\right\}$ denotes the corresponding feature maps extracted from the training set with the encoder $f_{\theta}$, which can be described as $f_{i}=f_{\theta}\left(x_{i}\right)$. $U=\left\{u_{1}, u_{2}, \ldots, u_{N}\right\}$ denotes the feature vectors got from the feature maps after the pooling operation. $u_{q}$ is the corresponding feature vector of the query instance $q$ extracted with encoder $f_{\theta}$. $\Phi=\left\{\phi_{1}, \phi_{2}, \ldots, \phi_{C}\right\}$ denotes  $C$ cluster representations in the training. Note that the number of the cluster $C$ can vary according to clustering results.  

Memory-based cluster contrast learning frameworks have achieved the state-of-the-art performance by taking advantage of memory mechanism and contrastive learning \cite{ge2020self, dai2021cluster}. Specifically, these methods utilize Kmeans \cite{macqueen1967some} or DBSCAN \cite{ester1996density} to generate pseudo labels for unlabeled samples. Thus a pseudo labeled dataset $X^{\prime}=\left\{\left(x_{1}, y_{1}\right),\left(x_{2}, y_{2}\right), \ldots,\left(x_{N}, y_{N^{\prime}}\right)\right\}$ can be obtained, where $y_{i} \in\{1, \ldots, C\}$ is the pseudo label generated for the $i-th$ sample and $N^{\prime}$ is the number of the labeled samples in the dataset. Then contrastive learning and memory mechanism can be applied on the pseudo labeled dataset. Among existing  memory-based cluster contrast learning frameworks, cluster contrast learning \cite{dai2021cluster} has achieved impressive performance by implementing contrastive learning on the cluster-level cluster memory dictionaries as following:
\begin{equation}\label{CCL}
L=-\log \frac{\exp \left(u_{q} \cdot \phi_{+} / \tau\right)}{\sum_{k=0}^{C} \exp \left(u_{q} \cdot \phi_{k} / \tau\right)},
\end{equation}
where $u_{q}$ is the feature vector of the query sample. $\phi_{k}$ is the centroid feature vector representing the $k-th$ cluster stored in the memory, which is initialized by the average feature vector of samples in the $k-th$ cluster and $\phi_{+}$ is the centroid feature vector of the cluster the query sample belongs to. $C$ is the number of the cluster. $\tau$ is the temperature hyper-parameter. Then the centroid feature vector stored in the memory dictionary sets can be updated in the following way:
\begin{equation}\label{eq2.5}
\phi_{k}=m \phi_{k}+(1-m) u_{q},
\end{equation}
where $m$ is the momentum updating factor and $k$ is the index of the cluster query sample belongs to.

\begin{figure*}[thbp] 
\centering
\includegraphics[width=1.0\textwidth,height=6.5cm]{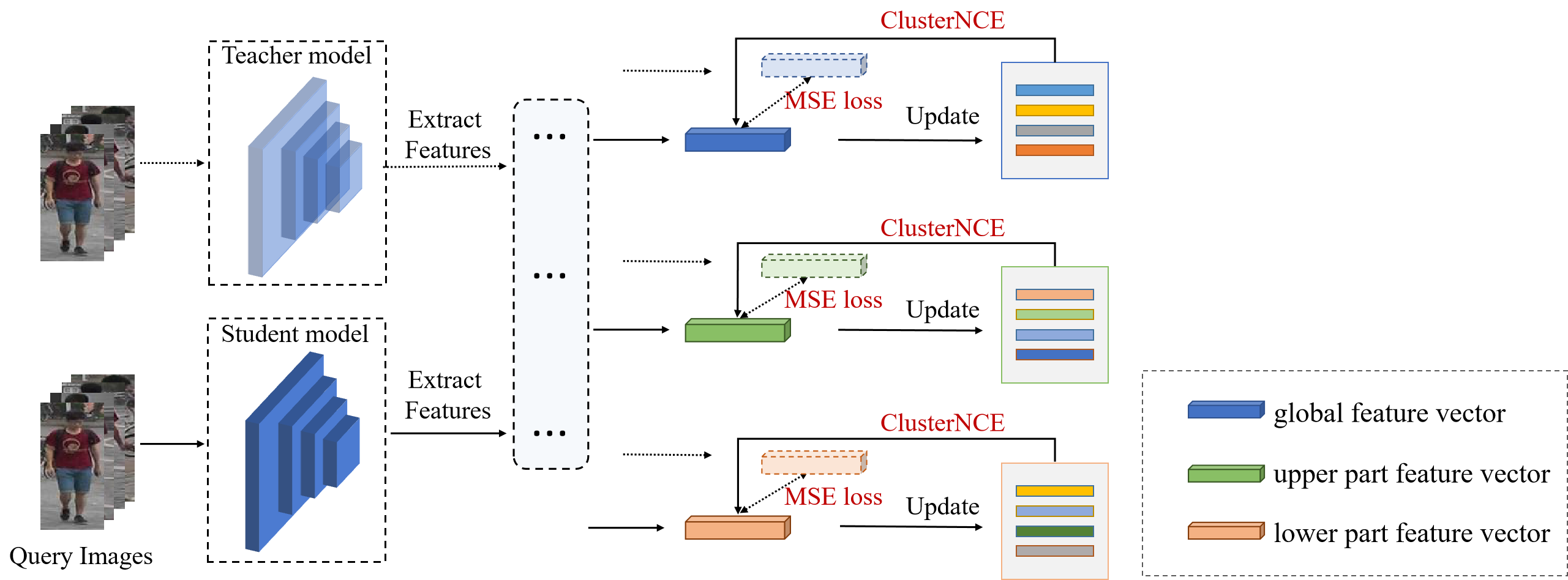}
\caption{Label noise purification module. We fixed the teacher model to learn the student model. The ClusterNCE loss and L2 loss are applied to update the student model. }\label{fig2b}
\end{figure*}

Our unsupervised person re-identification is implemented in the framework of the cluster contrast learning. To achieve the goal of purification in the unsupervised person re-identification. We design two functional components, namely FP module and LP module, as shown in Fig. \ref{fig2a} and Fig. \ref{fig2b}, respectively. The former takes into account the features from two local views to enrich the feature representation and purify the inherent feature bias of the global feature confront. Meanwhile, the latter aims to purify the label noise by taking advantage of the knowledge of teacher model in an offline scheme.

\subsection{Feature Purification Module}
\label{FP}
Although most works only utilize the global feature map for the unsupervised person ReID \cite{kumar2020unsupervised, ge2020structured, ge2019mutual, dai2021cluster}, the inherent feature bias of global feature may hinder the learning of the model for differing different persons as they tend to capture the most salient clues of the appearance while ignoring some detailed local cues. From this view, we propose the FP module, which aims to take advantage of extra local views to enhance the feature learning process by encouraging the model to discover more discriminative local cues in the feature representation. The procedure of the module is shown in Fig. \ref{fig2a}. To clearly describe the proposed FP module, we divide this module into three sub-modules as follows. 

\subsubsection{Multi-view features generation process}
Given an unlabeled training set $X=\left\{x_{1}, x_{2}, \ldots, x_{N}\right\}$, where $N$ is the number of the samples in the dataset. We can get the corresponding feature maps $F=\left\{f_{1}, f_{2}, \ldots, f_{N}\right\}$ with the encoder $f_{\theta}$. Then we split feature maps in $F$ into two parts  horizontally, which are denoted as $F^{{up}}=\left\{f_{1}^{{up}}, f_{2}^{{up}}, \ldots, f_{N}^{{up}}\right\}$ and $F^{ {dw}}=\left\{f_{1}^{{dw}}, f_{2}^{{dw}}, \ldots, f_{N}^{{dw}}\right\}$  respectively. To get the feature vectors from them, Generalized-Mean (GEM) pooling operations are applied on these feature branches independently. As a result, we can get three sets of feature vectors respectively.
\begin{equation}\label{eq1}
\left\{\begin{array}{l}
U^{  {gb}}=\left\{u_{1}^{  {gb}}, u_{2}^{  {gb}}, \ldots, u_{N}^{  {gb}}\right\} \\
U^{  {up}}=\left\{u_{1}^{  {up}}, u_{2}^{  {up}}, \ldots, u_{N}^{  {up}}\right\} \\
U^{  {dw}}=\left\{u_{1}^{  {dw}}, u_{2}^{  {dw}}, \ldots, u_{N}^{  {dw}}\right\}
\end{array},\right.
\end{equation}
where $U^{  {gb}}$, $U^{  {up}}$ and $U^{  {dw}}$ are three sets of feature vectors respectively. Compared with the global feature representations, feature vectors from these local views can introduce more detailed and complementary information about the person. Note that feature maps of introduced two local views are directly split from the global feature maps, thus the generation of these local feature vectors bring no extra computation burden to the model.    

\subsubsection{Pseudo labels generation process}
After getting the three sets of feature vectors in equation \ref{eq1}, following SPCL \cite{ge2020self} and CCL \cite{dai2021cluster}, we also apply DBSCAN \cite{ester1996density} clustering algorithm on these feature vectors to generate pseudo labels. Unlike these works which only utilize global features, we aim to generate pseudo labels by taking advantage of both global and local features. Our motivation is that as global features tend to capture the most salient cues, some non-salient but important detailed local cues can be easily ignored due to limited scales and less diversities of the training dataset. Thus images with different identities but similar appearance could be easily merged to the same cluster if we only utilize global features in the pseudo labels generation process. To mitigate such issue, we capture pairwise relations of samples by taking into consideration more detailed information. Specifically, with global and local feature vector sets $U^{  {gb}}$, $U^{  {up}}$ and $U^{  {dw}}$, the pairwise distance matrix of the dataset can be calculated independently, which are denoted as $D^{{gb}}$, $D^{  {up}}$ and $D^{ {dw}}$. Then a re-weighted pairwise distance matrix can be achieved using the following function:
\begin{equation}\label{eq1.5}
D=\left(1-2\lambda_{1}\right) D^{ {gb}}+\lambda_{1} D^{ {up}}+\lambda_{1} D^{ {dw}},
\end{equation}
where $D$ is the re-weighted pairwise distance matrix, $\lambda_{1}$ is the balancing factor. Then the pseudo labels $\tilde{Y}$ can be generated by DBSCAN clustering algorithm with matrix $D$. In this way, we can get a pseudo labeled dataset $X^{\prime}=\left\{\left(x_{1}, y_{1}\right),\left(x_{2}, y_{2}\right), \ldots,\left(x_{N}, y_{N^{\prime}}\right)\right\}$, where $N^{\prime}$ is the number of the pseudo labeled dataset. Note that $N^{\prime}$ is smaller than the number of samples in the original dataset $N$ due to the existence of outliers in the clustering  process. 

\subsubsection{Multi-view cluster centers generation process}
Following CCL \cite{dai2021cluster}, we also implement contrastive learning on the cluster-level memory dictionaries to avoid the problem of inconsistency in the memory updating process. Specifically, as the pseudo labeled dataset is obtained, then cluster centroids in the memory are initialized by the corresponding mean feature vectors and the pseudo labels as following:
\begin{equation}\label{eq2}
\phi_{k}=\frac{1}{\left|C_{k}\right|} \sum_{i \in C_{k}} u_{i},
\end{equation}
where $C_{k}$ denotes the $k-th$ cluster $|\cdot|$ denotes the number of the instances in the corresponding cluster and $u_{i}$ is the feature vector of the $i-th$ sample. To mitigate the limited representation of global features, we also maintain features from local views to promote the model to discover more detailed information in the learning process. As shown in Fig.~\ref{fig2a}, these three branches calculate the cluster center according to Eq. \eqref{eq2} independently with their own feature vectors and the shared pseudo label set $\tilde{Y}$. Thus, we can get three sets of cluster centroid representations as following:
\begin{equation}\label{eq3}
\left\{\begin{array}{l}
\Phi^{ {gb}}=\left\{\phi_{1}^{ {gb}}, \phi_{2}^{ {gb}}, \ldots, \phi_{C}^{ {gb}}\right\} \\
\Phi^{ {up }}=\left\{\phi_{1}^{ {up }}, \phi_{2}^{ {up }}, \ldots, \phi_{C}^{ {up }}\right\} \\
\Phi^{ {dw }}=\left\{\phi_{1}^{ {dw }}, \phi_{2}^{ {dw }}, \ldots, \phi_{C}^{ {dw }}\right\}
\end{array},\right.
\end{equation}
where $C$ is the number of the clusters, as these three branches share the same pseudo labels, thus the number of clusters in these three banches are the same. $\phi_{i}^{ {gb }}$, $\phi_{i}^{ {up }}$ and $\phi_{i}^{ {dw }}$ are the $i-th$ cluster centers in these three branches. Note that the pseudo labels generation operation is applied before each epoch, thus it may change in the training process according to clustering result before each epoch. 

\subsection{Label Noise Purification Module}
The training process of state-of-the-art unsupervised person ReID methods can be regarded as two stages. First, pseudo labels are generated by dividing the dataset into diverse clusters, then the model is trained with the pseudo labels. These two stages are conducted in an iterative scheme \cite{ge2020self, dai2021cluster}. However, the noise will be inevitably introduced in the convergence process as the model initialized with ImageNet pre-trained ResNet-50~\cite{he2016deep} performs poorly on these person ReID datasets at the beginning, which may accumulate label noise during the training process. To relieve the issue, we propose the LP module, which aims to utilize the knowledge of the teacher to help the student model relieve the influence of the label noise. For a fair comparison, we take the model trained on the same dataset as the teacher model and the new ImageNet pre-trained initialized model as the student model, thus the structures of these two models are the same and it requires no extra information. Our proposed LP module works based on the assumption that although the structure of the student model has no superiority over the teacher model, the trained teacher model performs better than the initialized student model on this task. Thus the teacher model can provide more accurate pseudo labels at the beginning and guide the student model to reduce the inference of noisy labels in the training phase through knowledge distillation. 

Specifically, the teacher model is trained with cluster contrast learning and the proposed FP module with the following objective:
\begin{equation}\label{eq4}
\left\{\begin{array}{l}
L_{q}^{ {gb}}=-\log \frac{\exp \left(u_{q}^{ {gb }} \cdot \phi_{+}^{ {gb }} / \tau\right)}{\sum_{k=0}^{C} \exp \left(u_{q}^{ {gb }} \cdot \phi_{k}^{ {gb }} / \tau\right)} \\
L_{q}^{ {up}}=-\log \frac{\exp \left(u_{q}^{ {up }} \cdot \phi_{+}^{ {up }} / \tau\right)}{\sum_{k=0}^{C} \exp \left(u_{q}^{ {up }} \cdot \phi_{k}^{ {up }} / \tau\right)} \\
L_{q}^{ {dw}}=-\log \frac{\exp \left(u_{q}^{ {dw }} \cdot \phi_{+}^{ {dw }} / \tau\right)}{\sum_{k=0}^{C} \exp \left(u_{q}^{ {dw }} \cdot \phi_{k}^{ {dw }} / \tau\right)}
\end{array},\right.
\end{equation}
where $u_{q}^{ {*}}$ is the feature vector of the query instance $q$ from the corresponding view. $\phi_{k}^{ {*}}$ is the centroid feature vector representing the $k-th$ cluster stored in the memory. $\phi_{+}^{ {*}}$ is the centroid feature vector representing the cluster query instance $q$ belongs to stored in the memory. $\tau$ is the temperature hyper-parameter and $C$ is the number of the cluster. Different from SPCL \cite{ge2020self} and CCL \cite{dai2021cluster} which only use global features in the training process, we also maintain feature representations from local views in the training phase to promote the model to discover more detailed information as following:
\begin{equation}\label{eq7}
L_{ {stage} 1}=\left(1-\lambda_{2}\right) L_{q}^{ {gb}}+\lambda_{2} (L_{q}^{ {up}} + L_{q}^{ {dw}}),
\end{equation}
where $\lambda_{2}$ is the loss weight to balance the importance between global and local features and more details about the training process of teacher model can refer to Sec. \ref{teacher-training}. 

When the trained teacher model is prepared, then LP module can be applied on the student model. This module includes two parts, the warm up part and the knowledge distillation part. More details about these parts can refer to Sec. \ref{student-training}. As the initialized student model performs poorly in the person ReID, the generated pseudo labels will contain numerous label noise in the early training period, thus may cause the feature representation biased. To tackle the issue, in the warm up part, we directly utilize  the trained teacher model to generate pseudo labels and use its feature vectors to initialize the cluster center representations as in Eq. \eqref{eq3}. Then the student is directly trained with the pseudo labels and fixed cluster center representations generated by the teacher model for a short period. In this way, the student model can learn the knowledge directly from the teacher model in a fast way to generate more accurate pseudo labels in the early period of the training phase.

In the remaining training phase, given the pseudo labeled dataset and memory center dictionaries as described in Sec. \ref{FP}, the student model computes the objective function with multi-view knowledge distillation as following:
\begin{equation}\label{eq9}
\left\{\begin{array}{l}
L_{S t u}^{ {gb }}=L_{q}^{ {gb }}+\mu\left|\frac{u_{q}^{ {gb }}}{\left\|u_{q}^{ {gb }}\right\|}-\frac{\tilde{u}_{q}^{ {gb }}}{\left\|\tilde{u}_{q}^{ {gb }}\right\|}\right|_{2}^{2} \\
L_{S t u}^{ {up }}=L_{q}^{ {up }}+\mu\left|\frac{u_{q}^{ {up }}}{\left\|u_{q}^{ {up }}\right\|}-\frac{\tilde{u}_{q}^{ {up }}}{\left\|\tilde{u}_{q}^{ {up }}\right\|}\right|_{2}^{2} \\
L_{S t u}^{ {dw} }=L_{q}^{ {dw }}+\mu\left|\frac{u_{q}^{ {dw }}}{\left\|u_{q}^{ {dw }}\right\|}-\frac{\tilde{u}_{q}^{ {dw }}}{\left\|\tilde{u}_{q}^{ {dw }}\right\|}\right|_{2}^{2}
\end{array},\right.
\end{equation}
where $L_{S t u}^{ {gb }}$, $L_{S t u}^{ {up }}$ and $L_{S t u}^{ {dw }}$ are the objective functions of three branches of the student model. $L_{q}^{ {gb }}$, $L_{q}^{ {up }}$ and $L_{q}^{ {dw }}$ are the ClusterNCE loss presented in Eq. \eqref{eq4}. $\mu$ is the balancing factor. $\{u_{q}^{ {gb }}, u_{q}^{ {up }},$  $u_{q}^{ {dw }}\}$ and $\{\tilde{u}_{q}^{ {gb }}, \tilde{u}_{q}^{ {up }}, \tilde{u}_{q}^{ {dw }}\}$ are the three feature vectors of query $q$ in the student model and teacher model respectively. Therefore, the final objective function of the student model is as following:
\begin{equation}\label{eq10}
L_{ {stage} 2}=\left(1-\lambda_{2}\right) L_{S t u}^{ {gb }}+\lambda_{2} (L_{S t u}^{ {up }}+ L_{S t u}^{ {dw }}),
\end{equation}
where $\lambda_{2}$ is the balancing factor, which is the same as in Eq. \eqref{eq7}. Then the cluster feature representations stored in the memory dictionary sets are updated similar with the teacher model in the following way:
\begin{equation}\label{eq8}
\left\{\begin{array}{l}
\phi_{k}^{ {gb }}=m \phi_{k}^{ {gb }}+(1-m) u_{q}^{ {gb }}  \\
\phi_{k}^{ {up }}=m \phi_{k_{2}}^{ {up }}+(1-m) u_{q}^{ {up }} \\
\phi_{k}^{ {dw }}=m \phi_{k_{3}}^{ {dw }}+(1-m) u_{q}^{ {dw }}
\end{array},\right.
\end{equation}
where $m$ is the momentum updating factor. $k$ is the index of the cluster query belongs to, which is the same in these three branches as they share the same pseudo label set. The details of the training procedure of the student model are described in Sec. \ref{student-training}. Note that the pseudo label generation process and training process are conducted iteratively until the model converges. In the test phase, we only adopt the global feature branch for computation efficiency.

\IncMargin{1em}
\begin{algorithm} \SetKwData{Left}{left}\SetKwData{This}{this}\SetKwData{Up}{up}\SetKwData{In}{in} \SetKwFunction{Union}{Union}\SetKwFunction{FindCompress}{FindCompress} \SetKwInOut{Input}{input}\SetKwInOut{Output}{output}\SetKwInOut{Require}{Require}
	
	\Require{Unlabeled training data $X$} 
	\Require{Initialize the encoder $f_{\theta}$ with ImageNet-pretrained ResNet-50}
	\Require{Temperature hyper-parameter $\tau$ for Eq. \eqref{eq4}}
	\Require{Balancing factors $\lambda_{1}$ and $\lambda_{2}$ for Eq. \eqref{eq1.5} and Eq. \eqref{eq7}}
	\Require{Momentum updating factor $m$ for Eq. \eqref{eq8}}
	 \BlankLine 

	 \For{$n$ \In $[1, $num\_epochs$]$}{ 
        Extract feature vector sets $\left\{U^{ {gb }}, U^{ {up }}, U^{ {dw} }\right\}$ from $X$ by $f_{\theta}$\;
        Clustering $\left\{U^{ {gb }}, U^{ {up }}, U^{ {dw} }\right\}$ into $C$ clusters with Eq. \eqref{eq1.5} and DBSCAN\;
        Initialize three memory dictionaries individually with Eq. \eqref{eq2} \;
	 	\For{$i$ \In $[1, $num\_iterations$]$}{
	 	Sample $P \times K$ query images from $X$\;
	 	Compute objective function with Eq. \eqref{eq7} \; 
	 	Update cluster feature representations with Eq. \eqref{eq8}\;
 	 	   }
 	 } 
 	 	  \caption{Training process of the teacher model}
 	 	  \label{algo_teacher} 
 	 \end{algorithm}
 \DecMargin{1em}

\subsection{Training Process}
\subsubsection{Training process of the teacher model}
\label{teacher-training}
The detailed training process of the teacher model is shown in Algorithm \ref{algo_teacher}. Given the unlabeled dataset $X$ and the encoder $f_{\theta}$ initialized with parameters of ResNet-50 pretrained on ImageNet~\cite{he2016deep}. For each epoch, we can get the corresponding feature map set $F^{{gb}}$ with the encoder $f_{\theta}$. Then we split feature maps $F^{{gb}}$ into two parts horizontally, which are denoted as $F^{{up}}$ and $F^{ {dw}}$ respectively. Then GEM pooling is applied to get the corresponding feature vector sets $U^{ {gb }}$, $U^{ {up }}$ and $U^{ {dw }}$, respectively. To get the pseudo labels $\tilde{Y}$ for samples in dataset $X$, Eq. \eqref{eq1.5} and DBSCAN algorithm are applied. Then, Eq. \eqref{eq2} is used to initialize memory dictionaries individually for these three branches. When the pseudo labels and memory dictionaries are prepared, we start to train the model. Specifically, in each iteration, we firstly sample $P \times K$ query images from $X$ to update parameters of the model according to Eq. \eqref{eq7}, and then update features stored in memory dictionaries as Eq. \eqref{eq8}.

\IncMargin{1em}
\begin{algorithm} \SetKwData{Left}{left}\SetKwData{This}{this}\SetKwData{Up}{up}\SetKwData{In}{in} \SetKwFunction{Union}{Union}\SetKwFunction{FindCompress}{FindCompress} \SetKwInOut{Input}{input}\SetKwInOut{Output}{output}\SetKwInOut{Require}{Require}
	
	\Require{Unlabeled training data $X$} 
	\Require{Initialize the encoder $f_{\theta}$ with ImageNet-pretrained ResNet-50}
	\Require{The teacher encoder $\tilde{f}_{\theta}$ trained on the unlabeled training data $X$ using Algorithm \ref{algo_teacher}}
	\Require{Balancing factors $\mu$ for Eq. \eqref{eq9}}
	 \BlankLine 

 	 \tcp{{warm up period}}
 	 Extract feature vector sets $\left\{U^{ {gb }}, U^{ {up }}, U^{ {dw } }\right\}$ from $X$ by $\tilde{f}_{\theta}$\;
 	 Clustering $\left\{U^{ {gb }}, U^{ {up }}, U^{ {dw } }\right\}$ into $C$ clusters  with Eq. \eqref{eq1.5} and DBSCAN\;
 	 Initialize three memory dictionaries individually with Eq. \eqref{eq2} \;
	 	\For{$i$ \In $[1, num\_iterations\times 2]$}{
	 	Sample $P \times K$ query images from $X$\;
	 	Compute objective function with Eq. \eqref{eq7} \; 
 	 	   }
     \tcp{knowledge distillation period}
	 \For{$n$ \In $[1, $num\_epochs$]$}{ 
        Extract feature vector sets $\left\{U^{ {gb }}, U^{ {up }}, U^{ {dw } }\right\}$ from $X$ by $f_{\theta}$\;
        Clustering $\left\{U^{ {gb }}, U^{ {up }}, U^{ {dw } }\right\}$ into $C$ clusters with Eq. \eqref{eq1.5} and DBSCAN\;
        Initialize three memory dictionaries individually with Eq. \eqref{eq2} \;
	 	\For{$i$ \In $[1, $num\_iterations$]$}{
	 	Sample $P \times K$ query images from $X$\;
	 	Compute objective function with Eq. \eqref{eq10} \; 
	 	Update cluster feature representations with $m$ and Eq. \eqref{eq8}\;
 	 	   }
 	 } 
 	 	  \caption{Training process of the student model}
 	 	  \label{algo_student} 
 	 \end{algorithm}
 \DecMargin{1em} 

\subsubsection{Training process of the student model}
\label{student-training}
The detailed training process of the student model is shown in Algorithm \ref{algo_student}. Besides the unlabeled dataset $X$ and the encoder $f_{\theta}$ initialized with parameters of ResNet-50 pretrained on ImageNet, the teacher model trained following Algorithm \ref{algo_teacher} is also required. The training process of the student model includes two parts, the warm-up part and the knowledge distillation part. 

In the warm-up part, we directly utilize the trained
teacher model to encode the dataset $X$ into feature map set $F^{{gb}}$, and then use horizontally split operation and GEM pooling to obtain corresponding feature vector sets $U^{ {gb }}$, $U^{ {up }}$ and $U^{ {dw }}$. Then Eq. \eqref{eq1.5} and DBSCAN algorithm are applied to get the pseudo labels and initialized memory dictionaries for each branch as described in the training process of the teacher model. Then $P \times K$ query images are sampled from $X$ to update parameters of the model according to Eq. \eqref{eq7} without updating the features stored in memory dictionaries. Note that in this part we aim to use fixed cluster centers stored in memory dictionaries from the teacher model to help the student model directly learn in a fast way to avoid label noise accumulation in the early period. Then in the knowledge distillation part, for each epoch the the training procedure of the student model is the same as the teacher model except that we update parameters of the student model according to Eq. \eqref{eq10} which contains the regularization of knowledge distillation from a global-local manner.

\section{Experiment}
\subsection{Datasets and Evaluation Protocol}
\label{datasets}
We conduct experiments on three public person Re-ID benchmarks, including Market-1501 \cite{zheng2015scalable}, DukeMTMC-reID \cite{ristani2016performance} and MSMT17 \cite{wei2018person}. Market-1501 dataset contains 32,668 images of 1,501 IDs captured by 6 different cameras. DukeMTMC-reID dataset is another large-scale person ReID dataset, which contains 36,441 images of 702 IDs captured by 8 different cameras. While MSMT17 dataset contains 126,441 images of 1,041 IDs captured by 15 different cameras. These datasets are widely used in the person ReID tasks and the details of these three datasets are summarized in Table \ref{tab:1}.

Following existing person ReID works \cite{zheng2015scalable, ge2020self, dai2021cluster, cheng2022hybrid}, we also adopt mean average precision (mAP) and Cumulated Matching Characteristics (CMC) as the evaluation metrics, and we report Top-1, Top-5, and Top-10 of the CMC evaluation metric in the paper. For fair comparisons, we don't adopt any post processing techniques in the evaluation period. As the setting of other purely unsupervised ReID works, we don't use any labeled data or other source domain datasets in the training process.

\begin{table}
\caption{Statistics of three person ReID datasets used in our experiments.}
\label{tab:1}       
\renewcommand{\arraystretch}{1.5}
\centering
\setlength{\tabcolsep}{1.2mm}{
\begin{tabular}{c|ccccc}
\hline
\multirow{2}*{{Datasets}}  & \multicolumn{1}{c}{\multirow{2}{*}{{Cameras}}} & \multicolumn{2}{|c}{{Training}} & \multicolumn{2}{|c}{{Testing}} \\
\cline { 3 - 6 } & & \multicolumn{1}{|c}{{IDs}} &\multicolumn{1}{c|}{{Images}} &{Query} &{Gallery}\\
\hline Market-1501 \cite{zheng2015scalable} & \multicolumn{1}{c|}{6} & \multicolumn{1}{c}{751} & \multicolumn{1}{c}{12,936} & \multicolumn{1}{|c}{3,368}
 & \multicolumn{1}{c}{19,732}\\
 \hline DukeMTMC-reID \cite{ristani2016performance} & \multicolumn{1}{c|}{8} & \multicolumn{1}{c}{702} & \multicolumn{1}{c}{16,522} & \multicolumn{1}{|c}{2,228}
 & \multicolumn{1}{c}{17,661}\\
  \hline MSMT17 \cite{wei2018person} & \multicolumn{1}{c|}{15} & \multicolumn{1}{c}{1,041} & \multicolumn{1}{c}{32,621} & \multicolumn{1}{|c}{11,659}
 & \multicolumn{1}{c}{51,027}\\ 
 \hline
\end{tabular}
}
\end{table}

\begin{table*}
\caption{Experimental results of our proposed method and state-of-the-art methods on Market-1501, DukeMTMC-reID, and MSMT17. The top three results are marked as \color[RGB]{200,0,0}{red}\color[RGB]{0,0,0}{,} \textcolor{blue}{blue} and \color[RGB]{0,120,0}{green}\color[RGB]{0,0,0}{, respectively.}}
\label{tab:2}       
\renewcommand{\arraystretch}{1.3}
\centering
\setlength{\tabcolsep}{2.5mm}{
\begin{tabular}{c|c|cccc|cccc|cccc}
\hline
\multirow{2}*{{Method}}  & \multicolumn{1}{c}{\multirow{2}{*}{{Reference}}} & \multicolumn{4}{|c}{{Market-1501}} & \multicolumn{4}{|c}{{DukeMTMC-reID}} & \multicolumn{4}{|c}{{MSMT17}}\\
\cline { 3 - 14 } & & \multicolumn{1}{c}{{mAP}} &\multicolumn{1}{c}{{R1}} &{R5} &{R10} & \multicolumn{1}{c}{{mAP}} &\multicolumn{1}{c}{{R1}} &{R5} &{R10} & \multicolumn{1}{c}{{mAP}} &\multicolumn{1}{c}{{R1}} &{R5} &{R10}\\
\hline
\multicolumn{6}{l}{{Unsupervised  Domain  Adaption}}\\

\hline ECN \cite{zhong2019invariance} & \multicolumn{1}{c|}{CVPR'19} & \multicolumn{1}{c}{43.0} & \multicolumn{1}{c}{75.1} & \multicolumn{1}{c}{87.6}
 & \multicolumn{1}{c}{91.6} & \multicolumn{1}{|c}{40.4} & \multicolumn{1}{c}{63.3} & \multicolumn{1}{c}{75.8}
 & \multicolumn{1}{c}{80.4} & \multicolumn{1}{|c}{10.2} & \multicolumn{1}{c}{30.2} & \multicolumn{1}{c}{41.5}
 & \multicolumn{1}{c}{46.8}\\
 \hline MMCL \cite{wang2020unsupervised} & \multicolumn{1}{c|}{CVPR'20} & \multicolumn{1}{c}{60.4} & \multicolumn{1}{c}{84.4} & \multicolumn{1}{c}{92.8} & \multicolumn{1}{c}{95.0} 
 & \multicolumn{1}{|c}{51.4} & \multicolumn{1}{c}{72.4} & \multicolumn{1}{c}{82.9} & \multicolumn{1}{c}{85.0} & \multicolumn{1}{|c}{16.2} & \multicolumn{1}{c}{43.6} & \multicolumn{1}{c}{54.3} & \multicolumn{1}{c}{58.9}\\
  \hline JVTC \cite{li2020joint} & \multicolumn{1}{c|}{ECCV'20} & \multicolumn{1}{c}{61.1} & \multicolumn{1}{c}{83.8} & \multicolumn{1}{c}{93.0} & \multicolumn{1}{c}{95.2} 
 & \multicolumn{1}{|c}{56.2} & \multicolumn{1}{c}{75.0} & \multicolumn{1}{c}{85.1} & \multicolumn{1}{c}{88.2} & \multicolumn{1}{|c}{20.3} & \multicolumn{1}{c}{45.4} & \multicolumn{1}{c}{58.4} & \multicolumn{1}{c}{64.3}\\
   \hline DG-Net++ \cite{zou2020joint} & \multicolumn{1}{c|}{ECCV'20} & \multicolumn{1}{c}{61.7} & \multicolumn{1}{c}{82.1} & \multicolumn{1}{c}{90.2} & \multicolumn{1}{c}{92.7} 
 & \multicolumn{1}{|c}{63.8} & \multicolumn{1}{c}{78.9} & \multicolumn{1}{c}{87.8} & \multicolumn{1}{c}{90.4} & \multicolumn{1}{|c}{22.1} & \multicolumn{1}{c}{48.8} & \multicolumn{1}{c}{60.9} & \multicolumn{1}{c}{65.9}\\
   \hline MMT \cite{ge2019mutual} & \multicolumn{1}{c|}{ICLR'20} & \multicolumn{1}{c}{71.2} & \multicolumn{1}{c}{87.7} & \multicolumn{1}{c}{94.9} & \multicolumn{1}{c}{96.9} 
 & \multicolumn{1}{|c}{65.1} & \multicolumn{1}{c}{78.0} & \multicolumn{1}{c}{88.8} & \multicolumn{1}{c}{92.5} & \multicolumn{1}{|c}{23.3} & \multicolumn{1}{c}{50.1} & \multicolumn{1}{c}{63.9} & \multicolumn{1}{c}{69.8}\\
   \hline DCML \cite{chen2020deep} & \multicolumn{1}{c|}{ECCV'20} & \multicolumn{1}{c}{72.6} & \multicolumn{1}{c}{87.9} & \multicolumn{1}{c}{95.0} & \multicolumn{1}{c}{96.7} 
 & \multicolumn{1}{|c}{63.3} & \multicolumn{1}{c}{79.1} & \multicolumn{1}{c}{87.2} & \multicolumn{1}{c}{89.4} & \multicolumn{1}{|c}{-} & \multicolumn{1}{c}{-} & \multicolumn{1}{c}{-} & \multicolumn{1}{c}{-}\\
   \hline MEB \cite{zhai2020multiple} & \multicolumn{1}{c|}{ECCV'20} & \multicolumn{1}{c}{76.0} & \multicolumn{1}{c}{89.9} & \multicolumn{1}{c}{96.0} & \multicolumn{1}{c}{97.5} 
 & \multicolumn{1}{|c}{66.1} & \multicolumn{1}{c}{79.6} & \multicolumn{1}{c}{88.3} & \multicolumn{1}{c}{92.2} & \multicolumn{1}{|c}{-} & \multicolumn{1}{c}{-} & \multicolumn{1}{c}{-} & \multicolumn{1}{c}{-}\\
   \hline SPCL \cite{ge2020self} & \multicolumn{1}{c|}{NeurIPS'20} & \multicolumn{1}{c}{76.7} & \multicolumn{1}{c}{90.3} & \multicolumn{1}{c}{96.2} & \multicolumn{1}{c}{97.7} 
 & \multicolumn{1}{|c}{68.8} & \multicolumn{1}{c}{82.9} & \multicolumn{1}{c}{{90.1}} & \multicolumn{1}{c}{92.5} & \multicolumn{1}{|c}{26.8} & \multicolumn{1}{c}{53.7} & \multicolumn{1}{c}{65.0} & \multicolumn{1}{c}{69.8}\\
\hline HCD \cite{zheng2021online} & \multicolumn{1}{c|}{ICCV'21} & \multicolumn{1}{c}{{80.2}} & \multicolumn{1}{c}{91.4} & \multicolumn{1}{c}{-} & \multicolumn{1}{c}{-} 
 & \multicolumn{1}{|c}{{71.2}} & \multicolumn{1}{c}{{83.1}} & \multicolumn{1}{c}{-} & \multicolumn{1}{c}{-} & \multicolumn{1}{|c}{{29.3}} & \multicolumn{1}{c}{{56.1}} & \multicolumn{1}{c}{-} & \multicolumn{1}{c}{-}\\

\hline \multicolumn{6}{l}{{Fully Unsupervised}}\\
\hline BUC \cite{lin2019bottom} & \multicolumn{1}{c|}{AAAI'19} & \multicolumn{1}{c}{29.6} & \multicolumn{1}{c}{61.9} & \multicolumn{1}{c}{73.5} & \multicolumn{1}{c}{78.2} 
 & \multicolumn{1}{|c}{22.1} & \multicolumn{1}{c}{40.4} & \multicolumn{1}{c}{52.5} & \multicolumn{1}{c}{58.2} & \multicolumn{1}{|c}{-} & \multicolumn{1}{c}{-} & \multicolumn{1}{c}{-} & \multicolumn{1}{c}{-}\\
 \hline SSL \cite{lin2020unsupervised} & \multicolumn{1}{c|}{CVPR'20} & \multicolumn{1}{c}{37.8} & \multicolumn{1}{c}{71.7} & \multicolumn{1}{c}{83.8} & \multicolumn{1}{c}{87.4}
 & \multicolumn{1}{|c}{28.6} & \multicolumn{1}{c}{52.5} & \multicolumn{1}{c}{63.5} & \multicolumn{1}{c}{68.9} & \multicolumn{1}{|c}{-} & \multicolumn{1}{c}{-} & \multicolumn{1}{c}{-} & \multicolumn{1}{c}{-}\\
 \hline JVTC \cite{li2020joint} & \multicolumn{1}{c|}{ECCV'20} & \multicolumn{1}{c}{41.8} & \multicolumn{1}{c}{72.9} & \multicolumn{1}{c}{84.2} & \multicolumn{1}{c}{88.7}
 & \multicolumn{1}{|c}{42.2} & \multicolumn{1}{c}{67.6} & \multicolumn{1}{c}{78.0} & \multicolumn{1}{c}{81.6} & \multicolumn{1}{|c}{15.1} & \multicolumn{1}{c}{39.0} & \multicolumn{1}{c}{50.9} & \multicolumn{1}{c}{56.8}\\
  \hline MMCL \cite{wang2020unsupervised} & \multicolumn{1}{c|}{CVPR'20} & \multicolumn{1}{c}{45.5} & \multicolumn{1}{c}{80.3} & \multicolumn{1}{c}{89.4} & \multicolumn{1}{c}{92.3}
 & \multicolumn{1}{|c}{40.2} & \multicolumn{1}{c}{65.2} & \multicolumn{1}{c}{75.9} & \multicolumn{1}{c}{80.0} & \multicolumn{1}{|c}{11.2} & \multicolumn{1}{c}{35.4} & \multicolumn{1}{c}{44.8} & \multicolumn{1}{c}{49.8}\\
  \hline HCT \cite{zeng2020hierarchical} & \multicolumn{1}{c|}{CVPR'20} & \multicolumn{1}{c}{56.4} & \multicolumn{1}{c}{80.0} & \multicolumn{1}{c}{91.6} & \multicolumn{1}{c}{95.2}
 & \multicolumn{1}{|c}{50.7} & \multicolumn{1}{c}{69.6} & \multicolumn{1}{c}{83.4} & \multicolumn{1}{c}{87.4} & \multicolumn{1}{|c}{-} & \multicolumn{1}{c}{-} & \multicolumn{1}{c}{-} & \multicolumn{1}{c}{-}\\
  \hline CycAs \cite{wang2020cycas} & \multicolumn{1}{c|}{ECCV'20} & \multicolumn{1}{c}{64.8} & \multicolumn{1}{c}{84.8} & \multicolumn{1}{c}{-} & \multicolumn{1}{c}{-}
 & \multicolumn{1}{|c}{60.1} & \multicolumn{1}{c}{77.9} & \multicolumn{1}{c}{-} & \multicolumn{1}{c}{-} & \multicolumn{1}{|c}{26.7} & \multicolumn{1}{c}{50.1} & \multicolumn{1}{c}{-} & \multicolumn{1}{c}{-}\\
  \hline GCL \cite{chen2021joint} & \multicolumn{1}{c|}{CVPR'21} & \multicolumn{1}{c}{66.8} & \multicolumn{1}{c}{87.3} & \multicolumn{1}{c}{93.5} & \multicolumn{1}{c}{95.5}
 & \multicolumn{1}{|c}{62.8} & \multicolumn{1}{c}{82.9} & \multicolumn{1}{c}{87.1} & \multicolumn{1}{c}{88.5} & \multicolumn{1}{|c}{21.3} & \multicolumn{1}{c}{45.7} & \multicolumn{1}{c}{58.6} & \multicolumn{1}{c}{64.5}\\
  \hline SPCL \cite{ge2020self} & \multicolumn{1}{c|}{NeurIPS'20} & \multicolumn{1}{c}{73.1} & \multicolumn{1}{c}{88.1} & \multicolumn{1}{c}{95.1} & \multicolumn{1}{c}{97.0}
 & \multicolumn{1}{|c}{65.3} & \multicolumn{1}{c}{81.2} & \multicolumn{1}{c}{90.3} & \multicolumn{1}{c}{92.2} & \multicolumn{1}{|c}{19.1} & \multicolumn{1}{c}{42.3} & \multicolumn{1}{c}{55.6} & \multicolumn{1}{c}{61.2}\\

   \hline HCD \cite{zheng2021online} & \multicolumn{1}{c|}{ICCV'21} & \multicolumn{1}{c}{78.1} & \multicolumn{1}{c}{91.1} & \multicolumn{1}{c}{96.4} & \multicolumn{1}{c}{97.7}
 & \multicolumn{1}{|c}{65.6} & \multicolumn{1}{c}{79.8} & \multicolumn{1}{c}{88.6} & \multicolumn{1}{c}{91.6} & \multicolumn{1}{|c}{26.9} & \multicolumn{1}{c}{53.7} & \multicolumn{1}{c}{65.3} & \multicolumn{1}{c}{70.2}\\
 \hline ICE \cite{chen2021ice} & \multicolumn{1}{c|}{ICCV'21} & \multicolumn{1}{c}{79.5} & \multicolumn{1}{c}{{92.0}} & \multicolumn{1}{c}{{97.0}} & \multicolumn{1}{c}{{98.1}} 
 & \multicolumn{1}{|c}{67.2} & \multicolumn{1}{c}{81.3} & \multicolumn{1}{c}{{90.1}} & \multicolumn{1}{c}{{93.0}} & \multicolumn{1}{|c}{{29.8}} & \multicolumn{1}{c}{{59.0}} & \multicolumn{1}{c}{{71.7}} & \multicolumn{1}{c}{{77.0}}\\
 
   \hline CCL \cite{dai2021cluster} & \multicolumn{1}{c|}{Arxiv'21} & \multicolumn{1}{c}{{82.6}} & \multicolumn{1}{c}{{93.0}} & \multicolumn{1}{c}{{97.0}} & \multicolumn{1}{c}{{98.1}}
 & \multicolumn{1}{|c}{\color[RGB]{0,120,0}{72.8}} & \multicolumn{1}{c}{\textcolor{blue}{85.7}} & \multicolumn{1}{c}{\color[RGB]{0,120,0}{92.0}} & \multicolumn{1}{c}{\color[RGB]{0,120,0}{93.5}} & \multicolumn{1}{|c}{{33.3}} & \multicolumn{1}{c}{{63.3}} & \multicolumn{1}{c}{{{73.7}}} & \multicolumn{1}{c}{{{77.8}}}\\
    \hline MCL \cite{jin2021meta} & \multicolumn{1}{c|}{Arxiv'21} & \multicolumn{1}{c}{{82.9}} & \multicolumn{1}{c}{{92.7}} & \multicolumn{1}{c}{\color[RGB]{0,120,0}{97.6}} & \multicolumn{1}{c}{\textcolor{blue}{98.7}}
 & \multicolumn{1}{|c}{{-}} & \multicolumn{1}{c}{{-}} & \multicolumn{1}{c}{{-}} & \multicolumn{1}{c}{{-}} & \multicolumn{1}{|c}{\textcolor{blue}{38.2}} & \multicolumn{1}{c}{\color[RGB]{0,120,0}{66.5}} & \multicolumn{1}{c}{\color[RGB]{0,120,0}{{75.2}}} & \multicolumn{1}{c}{\color[RGB]{0,120,0}{{79.7}}}\\
     \hline HDCPD \cite{cheng2022hybrid} & \multicolumn{1}{c|}{TIP'22} & \multicolumn{1}{c}{\color[RGB]{0,120,0}{84.5}} & \multicolumn{1}{c}{\color[RGB]{0,120,0}{93.5}} & \multicolumn{1}{c}{\color[RGB]{0,120,0}{97.6}} & \multicolumn{1}{c}{{98.6}}
 & \multicolumn{1}{|c}{\textcolor{blue}{73.5}} & \multicolumn{1}{c}{\color[RGB]{0,120,0}{85.4}} & \multicolumn{1}{c}{\textcolor{blue}{92.2}} & \multicolumn{1}{c}{\textcolor{blue}{94.5}} & \multicolumn{1}{|c}{{24.6}} & \multicolumn{1}{c}{{50.2}} & \multicolumn{1}{c}{{{61.4}}} & \multicolumn{1}{c}{{{65.7}}}\\
 \hline PPLR \cite{cho2022part} & \multicolumn{1}{c|}{CVPR'22} & \multicolumn{1}{c}{{81.5}} & \multicolumn{1}{c}{{92.8}} & \multicolumn{1}{c}{{97.1}} & \multicolumn{1}{c}{{98.1}}
 & \multicolumn{1}{|c}{{-}} & \multicolumn{1}{c}{{-}} & \multicolumn{1}{c}{{-}} & \multicolumn{1}{c}{{-}} & \multicolumn{1}{|c}{{31.4}} & \multicolumn{1}{c}{{61.1}} & \multicolumn{1}{c}{{{73.4}}} & \multicolumn{1}{c}{{{77.8}}}\\
 \hline ISE \cite{zhang2022implicit} & \multicolumn{1}{c|}{CVPR'22} & \multicolumn{1}{c}{\textcolor{blue}{85.3}} & \multicolumn{1}{c}{\textcolor{blue}{94.3}} & \multicolumn{1}{c}{\color[RGB]{200,0,0}{98.0}} & \multicolumn{1}{c}{\color[RGB]{200,0,0}{98.8}}
 & \multicolumn{1}{|c}{{-}} & \multicolumn{1}{c}{{-}} & \multicolumn{1}{c}{{-}} & \multicolumn{1}{c}{{-}} & \multicolumn{1}{|c}{\color[RGB]{0,120,0}{37.0}} & \multicolumn{1}{c}{\textcolor{blue}{67.6}} & \multicolumn{1}{c}{{\textcolor{blue}{77.5}}} & \multicolumn{1}{c}{{\textcolor{blue}{81.0}}}\\
   \hline Ours & \multicolumn{1}{c|}{-} & \multicolumn{1}{c}{\color[RGB]{200,0,0}{85.8}} & \multicolumn{1}{c}{\color[RGB]{200,0,0}{94.5}} & \multicolumn{1}{c}{\textcolor{blue}{97.8}} & \multicolumn{1}{c}{\textcolor{blue}{98.7}}
 & \multicolumn{1}{|c}{\color[RGB]{200,0,0}{76.2}} & \multicolumn{1}{c}{\color[RGB]{200,0,0}{86.7}} & \multicolumn{1}{c}{\color[RGB]{200,0,0}{93.0}} & \multicolumn{1}{c}{\color[RGB]{200,0,0}{94.3}} & \multicolumn{1}{|c}{\color[RGB]{200,0,0}{39.5}} & \multicolumn{1}{c}{\color[RGB]{200,0,0}{67.9}} & \multicolumn{1}{c}{\color[RGB]{200,0,0}{78.0}} & \multicolumn{1}{c}{\color[RGB]{200,0,0}{81.6}}\\
 \hline
\end{tabular}
}
\end{table*}

\subsection{Implementations Details}
We use the Resnet-50 \cite{he2016deep} initialized with the parameters pre-trained on the ImageNet \cite{deng2009imagenet} as the backbone encoder. Following existing cluster contrast framework \cite{dai2021cluster}, we remove all sub-module layers after layer-4 and add GEM pooling followed by batch normalization layer \cite{ioffe2015batch} and L2-normalization layer. During training, we use the DBSCAN \cite{ester1996density} as clustering algorithm to generate pseudo labels at the beginning of each epoch. During test phase, we only adopt the feature vector of the first global feature branch for computation efficiency.

For training, each mini-batch contains 256 images of 16 pseudo person identities, which are resized as $256 \times 128$. For input images, random horizontal flipping, padding, random cropping, and random erasing \cite{zhong2020random} are applied. To train our model, Adam optimizer with weight decay 5e-4 is adopted. We set the initial learning rate as 3.5e-4, and reduce it every 20 epochs for a total of 50 epochs. The balancing factor $\lambda_{1}$ in Eq. \eqref{eq1.5} is set to 0.2 while the balancing factor $\lambda_{2}$ in Eq. \eqref{eq7} is set to 0.15. The balancing factor $\mu$ in Eq. \eqref{eq9} is set to 1. For DBSCAN clustering algorithm, the minimal number of neighbours is set to 4 and the maximum distance $d$ is set to 0.6 for Market1501 and DukeMTMC-reID while 0.7 for MSMT17. 

\subsection{Comparison with State-of-the-Arts}
We compare our proposed method with the state-of-the-art unsupervised person ReID methods, including UDA person ReID and fully unsupervised person ReID. The result is shown in Table \ref{tab:2}. We first list the state-of-the-art UDA methods, including ECN \cite{zhong2019invariance}, MMCL \cite{wang2020unsupervised}, JVTC \cite{li2020joint}, DG-Net++ \cite{zou2020joint}, MMT \cite{ge2019mutual}, DCML \cite{chen2020deep}, MEB \cite{zhai2020multiple}, SPCL \cite{ge2020self} and HCD \cite{zheng2021online}. Although these methods leverage the knowledge of the source domain, our proposed method outperforms all of them on these three datasets. The reason is probably that the gap between source and target domains is large and it is hard to transfer the knowledge from source domain to the target domain. 

Compared with the state-of-the-art fully unsupervised person ReID methods, our proposed method also achieves better performance. These methods include BUC \cite{lin2019bottom}, SSL \cite{lin2020unsupervised}, JVTC \cite{li2020joint}, MMCL \cite{wang2020unsupervised}, HCT \cite{zeng2020hierarchical}, CycAs \cite{wang2020cycas}, GCL \cite{chen2021joint}, SPCL \cite{ge2020self}, HCD \cite{zheng2021online}, ICE \cite{chen2021ice}, CCL \cite{dai2021cluster}, MCL \cite{jin2021meta}, HDCPD \cite{cheng2022hybrid}, PPLR \cite{cho2022part} and ISE \cite{zhang2022implicit}. Specifically, as shown in Table \ref{tab:2}, our proposed method achieves 85.8/94.5 in mAP/rank-1 accuracy on Market-1501 and 76.2/86.7 in mAP/rank-1 on DukeMTMC-reID. On the MSMT17, our method achieves 39.5 in mAP and 67.9 in rank-1 accuracy. These results validate the superiority of our proposed method. As our proposed purification modules are established on the framework of CCL \cite{dai2021cluster}, our method outperforms CCL by 3.2\%, 3.4\% and 6.2\% in terms of mAP on Market-1501, DukeMTMC-ReID and MSMT17 datasets, respectively. Compared with Market-1501 and DukeMTMC-ReID, more gains can be achieved on the MSMT17 dataset. The reason is probably that more noise exist in MSMT17 dataset as it is more challenging compared with other datasets.

\begin{table}[ht]
\caption{Ablation study on Market-1501, DukeMTMC-reID, and MSMT17.}
\label{tab:3}       
\renewcommand{\arraystretch}{1.2}
\centering
\begin{tabular}{c|cccccc}
\hline
\multirow{2}*{{Method}}  & \multicolumn{2}{c}{{Market-1501}} & \multicolumn{2}{|c}{\small{{DukeMTMC-reID}}} & \multicolumn{2}{|c}{{MSMT17}} \\
\cline { 2 - 7 } & \multicolumn{1}{c}{{mAP}} &\multicolumn{1}{c|}{{R1}} &{mAP} &\multicolumn{1}{c|}{{R1}} & \multicolumn{1}{c}{{mAP}} &{R1} \\
\hline  \multicolumn{1}{c|}{{Baseline}} & \multicolumn{1}{c}{{82.8}} & \multicolumn{1}{c|}{{92.7}} & \multicolumn{1}{c}{{73.5}} & \multicolumn{1}{c|}{{85.3}} & \multicolumn{1}{c}{{34.2}} & \multicolumn{1}{c}{{64.2}} \\
\hline  \makecell[c]{{Baseline} \\ {+FP}} & \multicolumn{1}{c}{{84.4}} & \multicolumn{1}{c|}{{93.5}} & \multicolumn{1}{c}{{74.7}} & \multicolumn{1}{c|}{{86.2}} & \multicolumn{1}{c}{{35.2}} & \multicolumn{1}{c}{{64.5}} \\
\hline  \makecell[c]{ {Baseline} \\  {+LP}} & \multicolumn{1}{c}{ {84.7}} & \multicolumn{1}{c|}{ {93.6}} & \multicolumn{1}{c}{{75.0}} & \multicolumn{1}{c|}{ {86.5}} & \multicolumn{1}{c}{ {37.6}} & \multicolumn{1}{c}{ {67.6}} \\
\hline  \makecell[c]{ {Baseline} \\  {+FP+LP}} & \multicolumn{1}{c}{\textbf{{85.8}}} & \multicolumn{1}{c|}{\textbf{{94.5}}} & \multicolumn{1}{c}{\textbf{{76.2}}} & \multicolumn{1}{c|}{\textbf{{86.7}}} & \multicolumn{1}{c}{{\textbf{39.5}}} & \multicolumn{1}{c}{{\textbf{67.9}}} \\
\hline
\end{tabular}
\end{table}

\subsection{Ablation Studies}
In this section, we study the effectiveness of different components and hyper-parameters in our proposed method. As our work is implemented based on the CCL \cite{dai2021cluster}, the hyper-parameters introduced in our method include the balancing factors $\lambda_{1}$, $\lambda_{2}$ and $\mu$. The other hyper-parameters follow the setting of CCL.

\subsubsection{Different combinations of the components}
As our method is combined with two different purification modules, we conduct experiments on the three person ReID datasets described in Sec. \ref{datasets} versus different combinations of different modules. As our work is implemented based on CCL, we take CCL as baseline and our proposed modules include FP module and LP module. As shown in Table \ref{tab:3}, the first line means the result of CCL on different person ReID datasets. Compared with previous methods, CCL can achieve a good performance by taking advantage of contrastive learning and cluster center memory, but it is still limited by the feature bias and label noise as mentioned in the paper. The second line is the result of the combination of CCL and our proposed FP module, compared with the first line we can find that our proposed FP module can improve the baseline by 1.6\%, 1.2\% and 1.0\% in terms of mAP on Market-1501, DukeMTMC-ReID and MSMT17 datasets. The third line is the result of the combination of CCL and our proposed LP module, compared with the first line, the improvement of 1.9\%, 1.5\% and 3.4\% in terms of mAP can be achieved on these three datasets. The last line denotes the result of the combination of CCL and our proposed two purification modules. Compared with the first line, the improvement of 3.0\%, 2.7\% and 5.3\% in terms of mAP can be achieved on these datasets. The result shows that our proposed two purification modules can work in a mutual benefit way and the baseline with these two modules can achieve the best performance. Furthermore, compare the third line with the first line we can also find that the LP module is more effective on MSMT17 than the other two datasets. 
The reason is probably that compared with Market-1501 and DukeMTMC-reID, the MSMT17 dataset is more challenging which contains more occluded images. Thus the LP module can mitigate the severe side effect of label noise introduced in the clustering process. 

\begin{figure}[thbp] 
\centering
\includegraphics[width=0.5\textwidth,height=4.4cm]{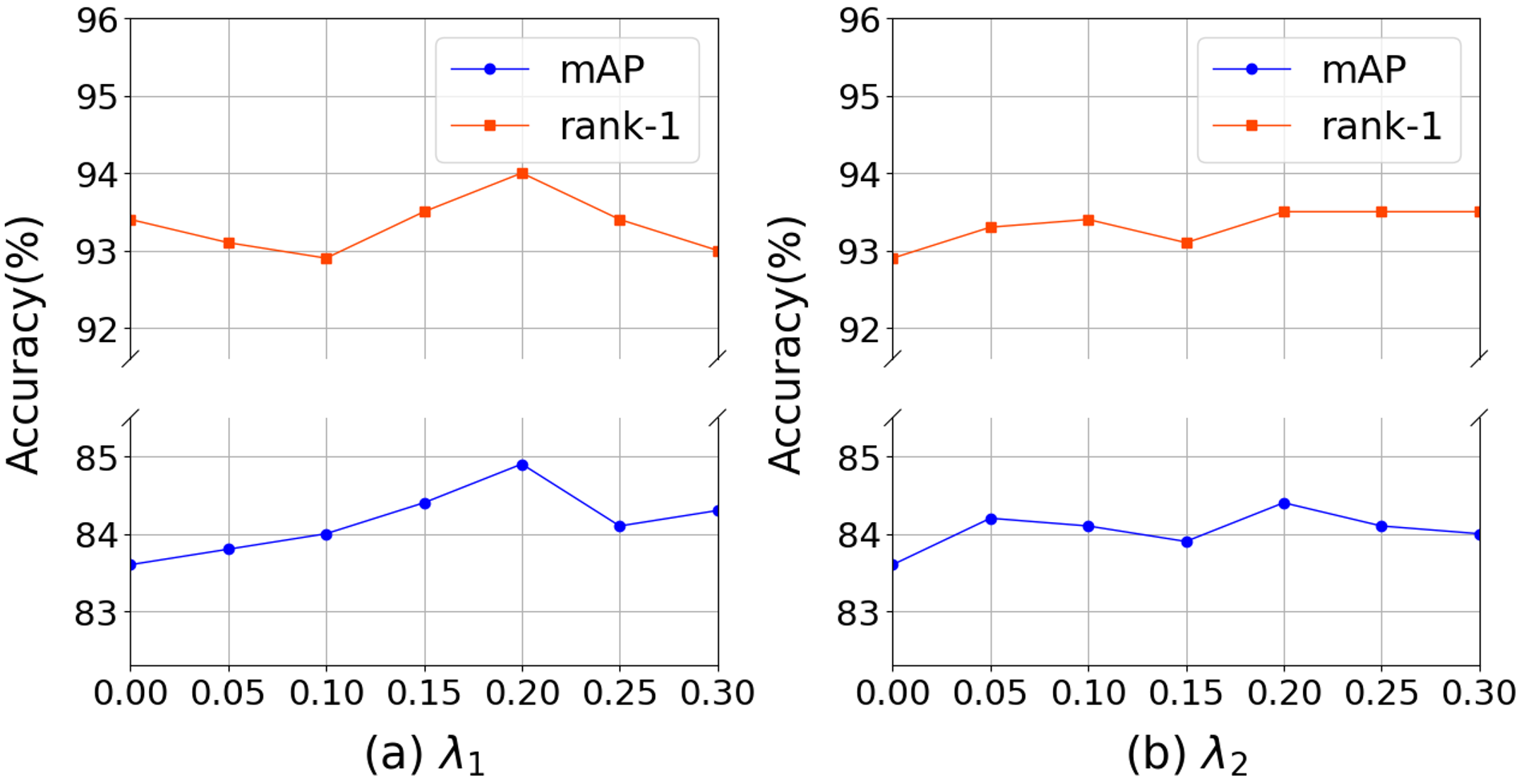}
\caption{Impact of hyper-parameter $\lambda_{1}$ and $\lambda_{2}$ of the teacher model on Market-1501. In (a) $\lambda_{2}$ is fixed to 0.15 while in (b) $\lambda_{1}$ is fixed to 0.2.}\label{fig5}
\end{figure}

\begin{figure}[thbp] 
\centering
\includegraphics[width=0.5\textwidth,height=4.4cm]{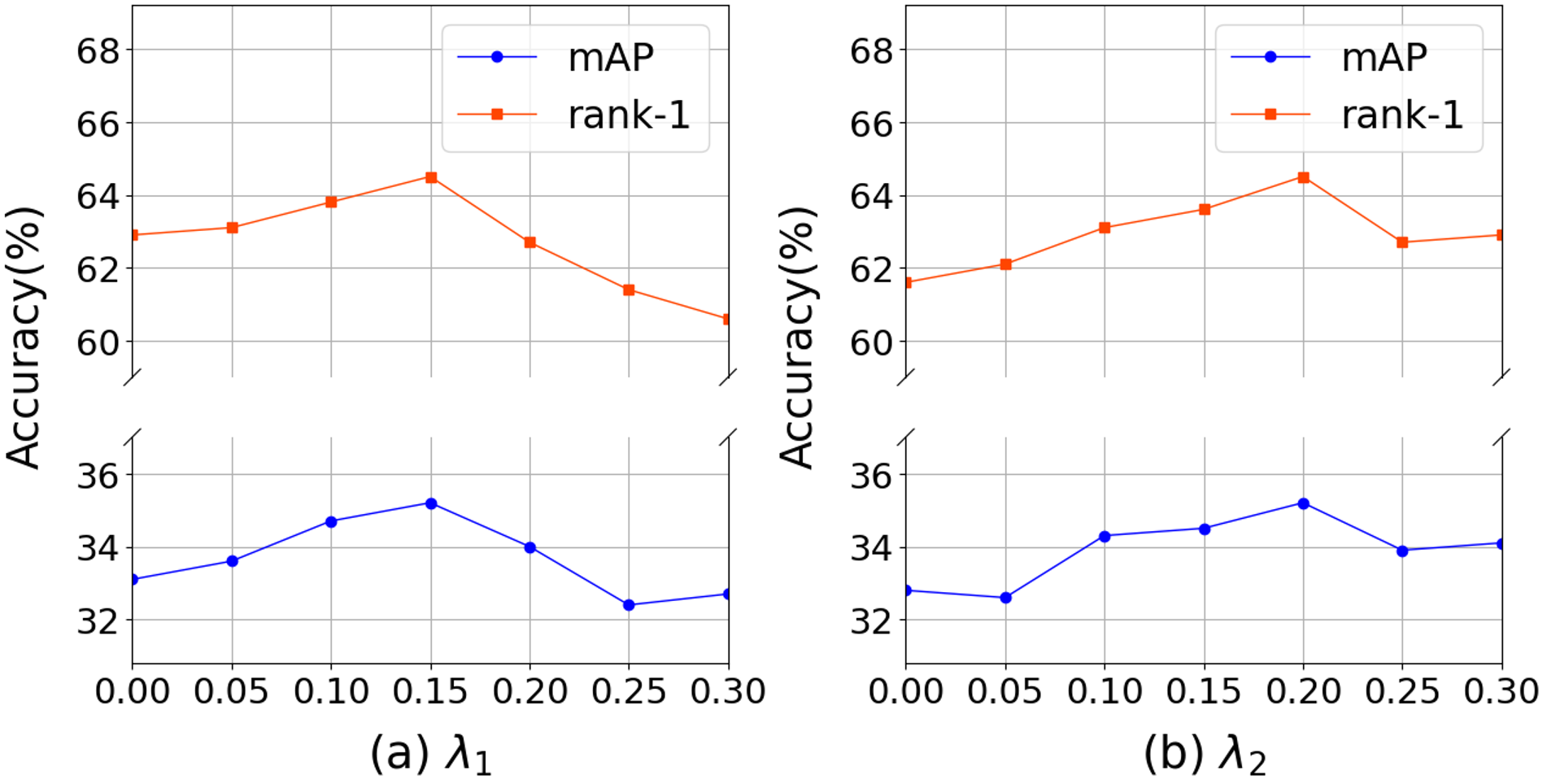}
\caption{Impact of hyper-parameter $\lambda_{1}$ and $\lambda_{2}$ of the teacher model on MSMT17. In (a) $\lambda_{2}$ is fixed to 0.15 while in (b) $\lambda_{1}$ is fixed to 0.2.}\label{fig5.5}
\end{figure}

\subsubsection{Balancing factors}
As the aim of our proposed extra two branches is to encourage the model to explore more discriminative local cues, the weights of these branches play an important role in our experiment. Fig. \ref{fig5} and Fig. \ref{fig5.5} report the result versus different values of balancing factors $\lambda_{1}$ and $\lambda_{2}$ in Eq. \eqref{eq1.5} and Eq. \eqref{eq7} on Market-1501 and MSMT17 datasets, respectively. Results in Fig. \ref{fig5} (a) and Fig. \ref{fig5.5} (a) show that the teacher model can achieve the bast results on Market-1501 and MSMT17 with $\lambda_{1}$ set to 0.2 and 0.15, respectively. Compared with $\lambda_{1}=0$, the above settings lead to better performance, which verifies the necessity of introducing local views into the pseudo label generation. To balance between different datasets, we set $\lambda_{1}$ to 0.15 by default. From Fig. \ref{fig5} (b) and Fig. \ref{fig5.5} (b), we can see that when $\lambda_{2}$ is set to 0.2, the teacher model can achieve the best results on both datasets. This setting leads to better performance compared with $\lambda_{2}=0$, which means the model can learn more discriminative feature representations with local features involved in the training process. 

Hyper-parameter $\mu$ in Eq. \eqref{eq9} is another important balancing factor, which determines the weight of the guidance of teacher in the overall training process. If $\mu$ is too small, then the student model will learn without enough guidance from the teacher model. On the other hand, if $\mu$ is too large, the student model will be forced to mimic the teacher model, which limits the generalization of the learned feature representations. Table \ref{tab:5} and Table \ref{tab:5.2} show the results under different values of $\mu$ on Market-1501 and MSMT17 datasets. As can be bseen, the model can achieve the best performance on both Market-1501 and MSMT17 datasets with $\mu$ set to 0.5 and 1.0, respectively. To balance between different datasets, we set $\mu$ to 1.0 in our experiments for all datasets. 
 
\begin{table}[ht]
\caption{Impact of hyper-parameter $\mu$ on Market-1501.}
\label{tab:5}       
\renewcommand{\arraystretch}{1.3}
\centering
\setlength{\tabcolsep}{5.0mm}{
\begin{tabular}{c|cccc}
\hline
\multirow{2}*{{$\mu$}}  & \multicolumn{4}{c}{{Market-1501}}  \\
\cline { 2 - 5 } & \multicolumn{1}{c}{{mAP}} &\multicolumn{1}{c}{{R1}} &{R5} &\multicolumn{1}{c}{{R10}}  \\
\hline  \multicolumn{1}{c|}{{0.5}} & \multicolumn{1}{c}{\textbf{86.1}} & \multicolumn{1}{c}{{94.3}} & \multicolumn{1}{c}{\textbf{98.0}} & \multicolumn{1}{c}{\textbf{98.7}} \\
\hline  \multicolumn{1}{c|}{{1.0}} & \multicolumn{1}{c}{{85.8}} & \multicolumn{1}{c}{\textbf{94.5}} & \multicolumn{1}{c}{{97.8}} & \multicolumn{1}{c}{\textbf{98.7}} \\
\hline  \multicolumn{1}{c|}{{1.5}} & \multicolumn{1}{c}{{85.8}} & \multicolumn{1}{c}{{93.9}} & \multicolumn{1}{c}{{97.4}} & \multicolumn{1}{c}{{98.4}} \\
\hline  \multicolumn{1}{c|}{{2.0}} & \multicolumn{1}{c}{{85.8}} & \multicolumn{1}{c}{{94.0}} & \multicolumn{1}{c}{{97.5}} & \multicolumn{1}{c}{{98.5}} \\
\hline  \multicolumn{1}{c|}{{2.5}} & \multicolumn{1}{c}{{85.9}} & \multicolumn{1}{c}{{94.4}} & \multicolumn{1}{c}{{97.8}} & \multicolumn{1}{c}{{98.3}} \\
\hline  \multicolumn{1}{c|}{{3.0}} & \multicolumn{1}{c}{{85.8}} & \multicolumn{1}{c}{{94.3}} & \multicolumn{1}{c}{{97.7}} & \multicolumn{1}{c}{{98.4}} \\
\hline
\end{tabular}
}
\end{table}

\begin{table}[ht]
\caption{Impact of hyper-parameter $\mu$ on MSMT17.}
\label{tab:5.2}       
\renewcommand{\arraystretch}{1.3}
\centering
\setlength{\tabcolsep}{5.0mm}{
\begin{tabular}{c|cccc}
\hline
\multirow{2}*{{$\mu$}}  & \multicolumn{4}{c}{{MSMT17}}  \\
\cline { 2 - 5 } & \multicolumn{1}{c}{{mAP}} &\multicolumn{1}{c}{{R1}} &{R5} &\multicolumn{1}{c}{{R10}}  \\
\hline  \multicolumn{1}{c|}{{0.5}} & \multicolumn{1}{c}{{39.3}} & \multicolumn{1}{c}{{67.5}} & \multicolumn{1}{c}{{77.6}} & \multicolumn{1}{c}{{81.4}} \\
\hline  \multicolumn{1}{c|}{{1.0}} & \multicolumn{1}{c}{\textbf{39.5}} & \multicolumn{1}{c}{\textbf{67.9}} & \multicolumn{1}{c}{\textbf{78.0}} & \multicolumn{1}{c}{\textbf{81.6}} \\
\hline  \multicolumn{1}{c|}{{1.5}} & \multicolumn{1}{c}{{38.8}} & \multicolumn{1}{c}{{67.4}} & \multicolumn{1}{c}{{77.2}} & \multicolumn{1}{c}{{81.1}} \\
\hline  \multicolumn{1}{c|}{{2.0}} & \multicolumn{1}{c}{{38.7}} & \multicolumn{1}{c}{{67.6}} & \multicolumn{1}{c}{{77.2}} & \multicolumn{1}{c}{{80.8}} \\
\hline  \multicolumn{1}{c|}{{2.5}} & \multicolumn{1}{c}{{37.9}} & \multicolumn{1}{c}{{66.8}} & \multicolumn{1}{c}{{76.5}} & \multicolumn{1}{c}{{80.2}} \\
\hline  \multicolumn{1}{c|}{{3.0}} & \multicolumn{1}{c}{{37.5}} & \multicolumn{1}{c}{{66.7}} & \multicolumn{1}{c}{{76.4}} & \multicolumn{1}{c}{{79.9}} \\
\hline
\end{tabular}
}
\end{table}

\begin{figure}[thbp] 
\centering
\includegraphics[width=0.4\textwidth,height=5.0cm]{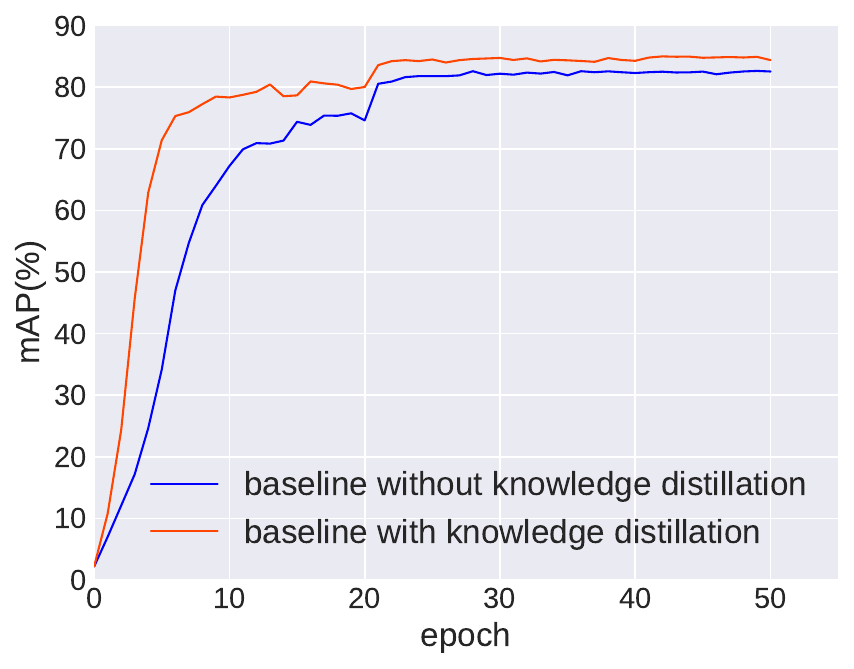}
\caption{Accuracy of our model with/without knowledge distillation during training on Market-1501. }\label{fig6}
\end{figure}

\begin{figure}[thbp] 
\centering
\includegraphics[width=0.4\textwidth,height=5.0cm]{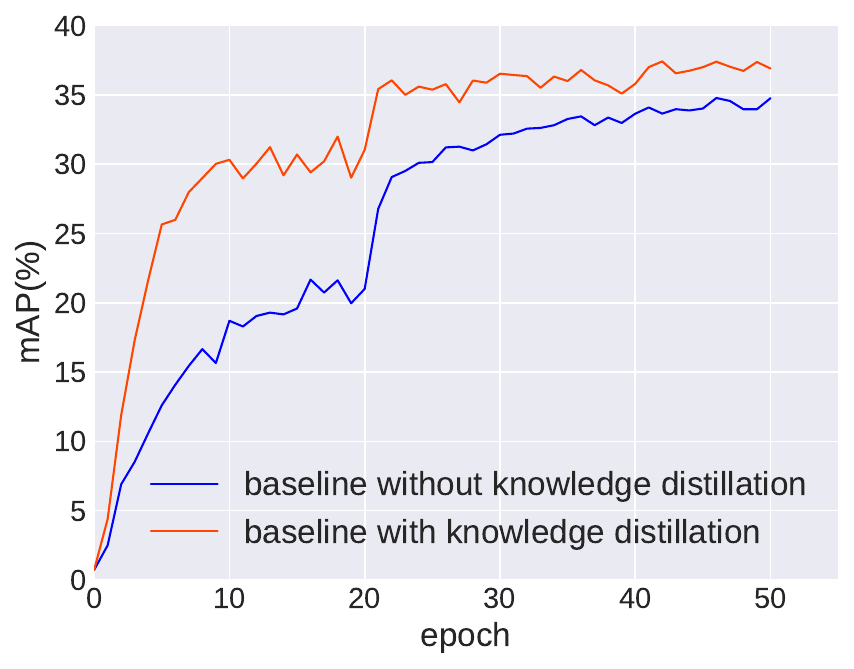}
\caption{Accuracy of our model with/without knowledge distillation during training on MSMT17. }\label{fig6.2}
\end{figure}

\subsubsection{Influence of knowledge distillation}
To investigate the influence of knowledge distillation, we show the test accuracy of the baseline with/without knowledge distillation in each epoch. Due to the lack of ground truth labels, the model has to be trained with the pseudo labels generated by clustering algorithm. In this way, the noise will be inevitably introduced in the convergence process as the model initialized with ImageNet pre-trained ResNet-50 performs poorly on these person ReID datasets. As shown in Fig. \ref{fig6} and Fig. \ref{fig6.2}, the student model with the offline knowledge distillation converges faster than its counterpart without knowledge distillation, since it mitigates the interference of noisy labels.

\begin{table}[htbp]
\caption{Results of different methods with IBN-ResNet-50 backbone on Market-1501, DukeMTMC-reID and MSMT17.}
\label{tab:7}       
\renewcommand{\arraystretch}{1.4}
\centering
\setlength{\tabcolsep}{2.1mm}{
\begin{tabular}{c|c|cccc}
\hline
{{dataset}}  & {method} & {{mAP}} &\multicolumn{1}{c}{{R1}} &{R5} &\multicolumn{1}{c}{{R10}}  \\
\hline 
\multirow{3}*{{Market-1501}} &  \multicolumn{1}{c|}{SPCL \cite{ge2020self}} &73.8 & 88.4 & 95.3 & 97.3 \\
& \multicolumn{1}{c|}{CCL \cite{dai2021cluster}} &84.1 & 93.2 & 97.6 & 98.2  \\
 & \multicolumn{1}{c|}{{Ours}} &\textbf{86.9} &\textbf{94.4} &\textbf{97.7} &\textbf{98.5}  \\
\hline

\multirow{3}*{{DukeMTMC-reID}} &  \multicolumn{1}{c|}{SPCL \cite{ge2020self}} &66.7 & 82.1 & 90.0 & 92.4 \\
&  \multicolumn{1}{c|}{CCL \cite{dai2021cluster}} &74.2 &85.8 &92.1 &94.2 \\
 & \multicolumn{1}{c|}{{Ours}} &\textbf{76.8} &\textbf{88.2} &\textbf{92.9} &\textbf{94.7}  \\

\hline 
\multirow{3}*{{MSMT17}} &  \multicolumn{1}{c|}{SPCL \cite{ge2020self}} &24.0 & 48.9 & 61.8 & 67.1 \\
&  \multicolumn{1}{c|}{CCL \cite{dai2021cluster}} &41.1 &69.1 &79.3 &83.1 \\
 & \multicolumn{1}{c|}{{Ours}} &\textbf{47.9} &\textbf{74.5} &\textbf{83.8} &\textbf{86.7}  \\
\hline

\end{tabular}
}
\end{table}

\begin{figure*}[thbp] 
\centering
\includegraphics[width=0.85\textwidth,height=8.5cm]{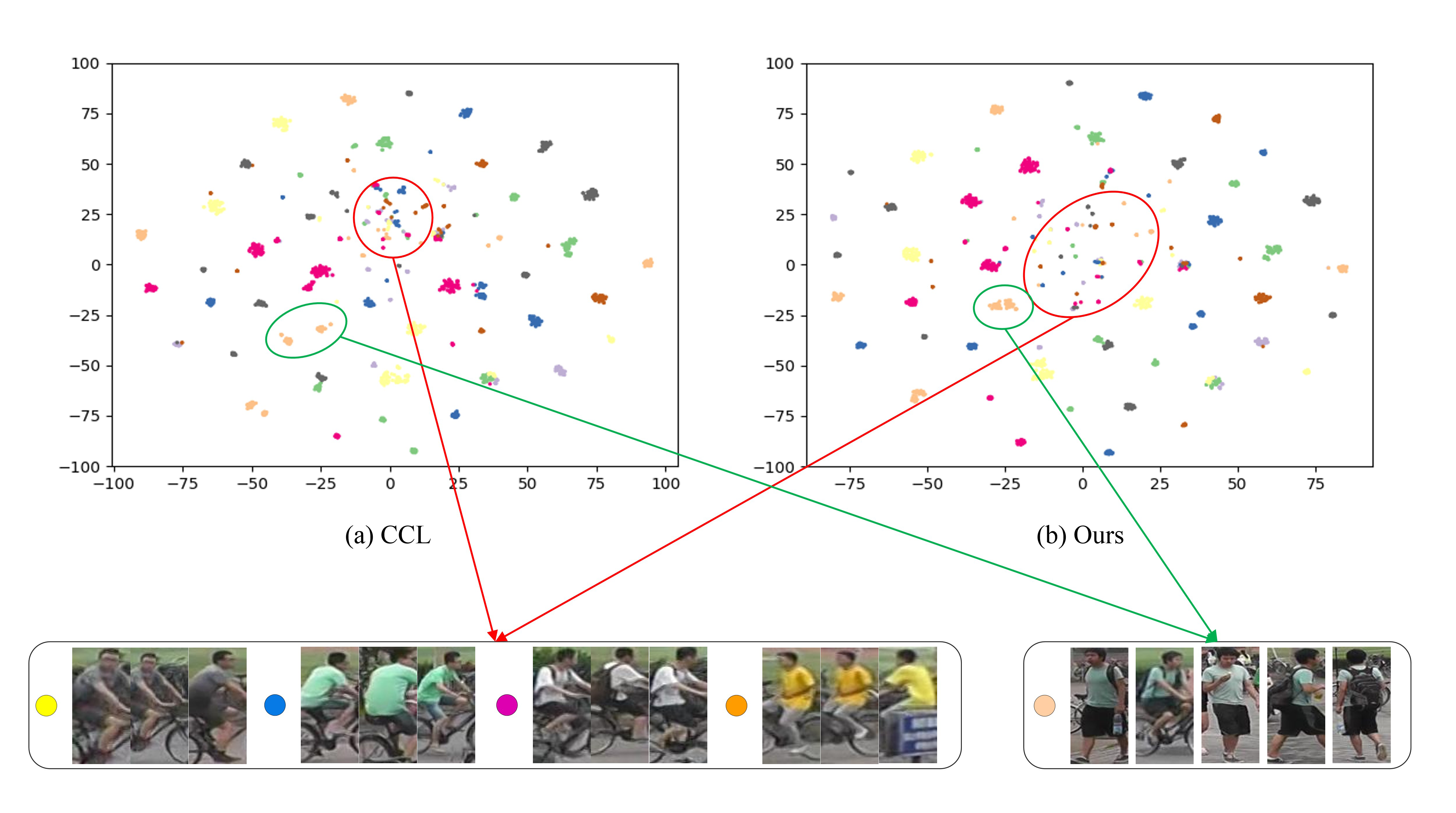}
\caption{T-SNE visualization of the learned features on a subset of Market-1501 training set (100 identities, 1,747 images). Points of the same color represent features of the same identity. We show detailed images of some points in the figure. The red circle contains some hard negative samples, which share similar appearance but have different identities, while the green circle contains hard positive samples, which have the same identity but quite different appearance. Compared with CCL, our method is more discriminative for the hard negative samples while have more compact features for the hard positive samples.}\label{fig10}
\end{figure*}

\begin{figure*}[thbp] 
\centering
\includegraphics[width=1.0\textwidth,height=10.5cm]{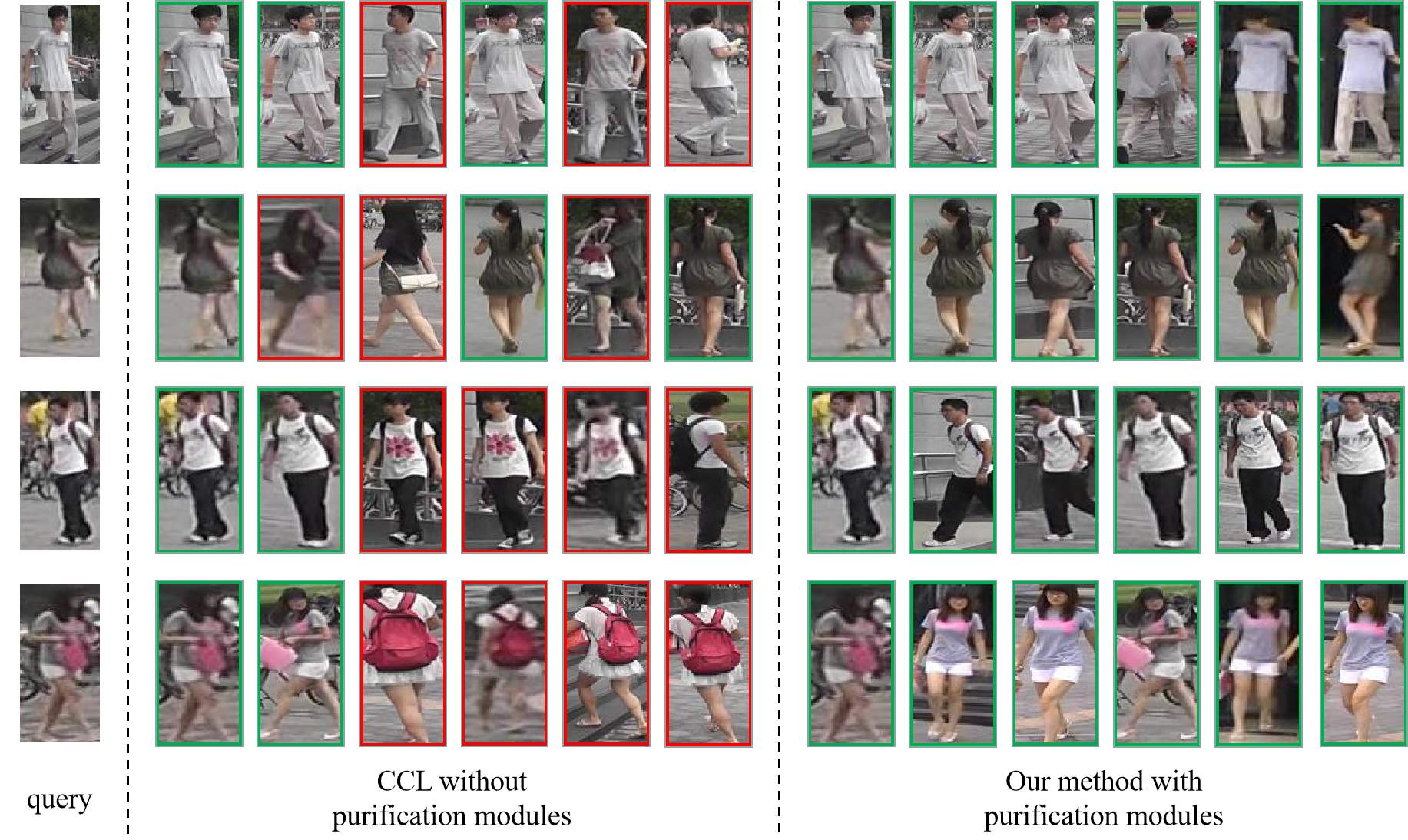}
\caption{Top 6 retrieval results of some hard queries on Market-1501 dataset. Note that the green/red boxes denote true/false retrieval results, respectively. }\label{visual}
\end{figure*}

\subsubsection{Compared with IBN-ResNet-50 and ResNet-50 backbones}
As Instance Normalization (IN) can learn features that are invariant to appearance changes, while Batch Normalization (BN) is essential for preserving content related information, IBN-Net \cite{pan2018two} can achieve better performance by integrating Instance Normalization and Batch Normalization. Thus, IBN-ResNet-50 can be regarded as a stronger baseline by replacing the BN in ResNet-50 with IBN. Comparing the results in Table \ref{tab:7} and Table~\ref{tab:2}, we can find that our proposed method can achieve better performance with the IBN-ResNet-50 backbone than the ResNet-50 backbone. 

\subsubsection{Qualitative analysis of visualization}
To further understand the discrimination ability of our method, we utilize t-SNE \cite{van2008visualizing} to visualize the features learned by the baseline and our method. As shown in Fig. \ref{fig10}, features of the same identity are usually clustered together in both CCL and our proposed method, which verifies the effectiveness of CCL and our method. More specifically, we show some detailed images of some points in the figure. The red circle contains some hard negative samples, which share similar appearance but have different identities, while the green circle contains hard positive samples, which have the same identity but quite different appearance. We can see that CCL cannot effectively distinguish these hard negative samples, i.e., they are close to each other in the embedding space and easily result in wrong clusters in the clustering process, while our method can distinguish them effectively, i.e., being more discriminative for the hard negative samples. On the other hand, for those hard positive samples, our method can produce more compact features than CCL.

We also present some retrieval examples with top 6 retrieved images in Fig. \ref{visual}. Our purification modules can greatly improve the performance of the baseline CCL. In the first two rows of Fig. \ref{visual}, the baseline gets some false results for the query due to the high similarity between different persons in terms of gender, clothes, etc., except the hair style. However, the baseline with our proposed purification modules can find the true retrieval results, and the reason is that our proposed FP module helps the baseline learn a more discriminative feature representation by capturing more detailed local cues, i.e., the hair style. Similarly, in the third row, the baseline with the proposed purification modules gets more accurate retrieval results by extracting more information about the logo of the shirt. In the last row, the baseline could be mislead by the similarity between the backpack and the blurry handbag. As a result, these images may be easily merged to the same cluster in the early period of pseudo label generation process and the model could be biased due to the noise accumulation. But our method can deliver true retrieval results, and the reason is that our proposed LP module can guide the student model to learn a more robust feature representation by taking advantage of the trained teacher model.


\section{Conclusion}
In the paper we propose the purification method for unsupervised person ReID. Two novel purification modules are devised. Specifically, the feature purification module takes into account the features from two local views to enrich the feature representation to purify the inherent feature bias of the global feature involved. The label noise purification module helps purify the label noise by taking advantage of the knowledge of teacher model in an offline scheme. Extensive experiments on three challenging person ReID datasets demonstrate the superiority of our method over state-of-the-art methods.

\ifCLASSOPTIONcaptionsoff
  \newpage
\fi

\end{document}